\documentclass[conference]{IEEEtran}
\usepackage{textgreek}
\usepackage[utf8]{inputenc}

\usepackage{bm}
\usepackage{amssymb}
\usepackage{amsmath}
\usepackage{mathtools}
\usepackage{nicefrac}  
\usepackage{cancel}
\makeatletter
\newcommand*{\T}{{\mathpalette\@transpose{}} }
\newcommand*{\@transpose}[2]{\raisebox{\depth}{$\m@th#1\intercal$}}
\makeatother
\DeclarePairedDelimiterX{\norm}[1]{\lVert}{\rVert_{2}}{#1}

\newcommand{\real}{\mathbb{R}}

\usepackage[linesnumbered,ruled,vlined]{algorithm2e}
\usepackage{algpseudocode}

\usepackage{booktabs}
\usepackage{multicol}
\usepackage{multirow}
\usepackage{tabularx}
\usepackage{etoolbox}
\newrobustcmd{\B}{\bfseries}  
\usepackage{csvsimple}
\usepackage{siunitx}  
\usepackage{diagbox}
\usepackage{pifont}  
\usepackage{todo}  
\usepackage[table]{xcolor}
\definecolor{red}{RGB}{200,50,50}
\definecolor{green}{RGB}{50,200,50}
\definecolor{blue}{RGB}{50,50,200}
\definecolor{orange}{RGB}{243,147,82}
\definecolor{purple}{RGB}{150,100,250}
\definecolor{yellow}{RGB}{231,198,100}
\definecolor{gray}{RGB}{98,95,75}
\definecolor{white}{RGB}{255,255,255}
\usepackage{lipsum}
\setlipsum{%
  par-before = \begingroup\color{gray},
  par-after = \endgroup
}
\usepackage{times}
\newcommand{\topic}[1]{\textbf{#1}\quad}
\usepackage{tcolorbox}  

\newtcolorbox{questionbox}{
    colback=white,
    colframe=red,
    arc=4mm, 
    boxrule=0.5pt,
    left=5pt,
    right=5pt,
    top=5pt,
    bottom=5pt,
    fonttitle=\bfseries,
    title={\centering Question}
}
\newtcolorbox{answerbox}{
    colback=white,
    colframe=black,
    sharp corners,
    boxrule=0.5pt,
    left=5pt,
    right=5pt,
    top=5pt,
    bottom=5pt,
    fonttitle=\bfseries,
    title={\centering Answer}
}

\usepackage{soul}

\makeatletter
\let\NAT@parse\undefined
\makeatother
\usepackage[numbers,sort&compress,square,comma]{natbib}  
\usepackage[bookmarks=true]{hyperref}
\hypersetup{colorlinks=true, urlcolor=blue}
\usepackage{cleveref}  
\crefname{lemma}{Lemma}{Lemmas}
\crefname{proposition}{Proposition}{Propositions}
\crefname{definition}{Definition}{Definitions}
\crefname{theorem}{Theorem}{Theorems}
\crefname{conjecture}{Conjecture}{Conjectures}
\crefname{corollary}{Corollary}{Corollaries}
\crefname{section}{Section}{Sections}
\crefname{appendix}{Appendix}{Appendices}
\crefname{figure}{Fig.}{Figs.}
\crefname{equation}{Eq.}{Eqs.}
\crefname{table}{Table}{Tables}
\crefname{algocf}{Algorithm}{Algorithms}
\crefname{problem}{Problem}{Problems}
\usepackage{balance}  
\apptocmd{\sloppy}{\hbadness 10000\relax}{}{}

\usepackage{graphicx}
\usepackage[caption=false,font=footnotesize]{subfig}
\usepackage{placeins}  
\usepackage{wrapfig}  
\usepackage{tikz}
\usetikzlibrary{positioning,shapes,bayesnet,calc,backgrounds}

\makeatletter
\newif\if@showgrid@grid
\newif\if@showgrid@left
\newif\if@showgrid@right
\newif\if@showgrid@below
\newif\if@showgrid@above
\tikzset{%
    every show grid/.style={},
    show grid/.style={execute at end picture={\@showgrid{grid=true,#1}}},%
    show grid/.default={true},
    show grid/.cd,
    labels/.style={font={\sffamily\small},help lines},
    xlabels/.style={},
    ylabels/.style={},
    keep bb/.code={\useasboundingbox (current bounding box.south west) rectangle (current bounding box.north west);},
    true/.style={left,below},
    false/.style={left=false,right=false,above=false,below=false,grid=false},
    none/.style={left=false,right=false,above=false,below=false},
    all/.style={left=true,right=true,above=true,below=true},
    grid/.is if=@showgrid@grid,
    left/.is if=@showgrid@left,
    right/.is if=@showgrid@right,
    below/.is if=@showgrid@below,
    above/.is if=@showgrid@above,
    false,
}
\def\@showgrid#1{%
    \begin{scope}[every show grid,show grid/.cd,#1]
        \if@showgrid@grid
            \begin{pgfonlayer}{background}
                \draw [help lines]
                (current bounding box.south west) grid
                (current bounding box.north east);
                \pgfpointxy{1}{1}%
                \edef\xs{\the\pgf@x}%
                \edef\ys{\the\pgf@y}%
                \pgfpointanchor{current bounding box}{south west}
                \edef\xa{\the\pgf@x}%
                \edef\ya{\the\pgf@y}%
                \pgfpointanchor{current bounding box}{north east}
                \edef\xb{\the\pgf@x}%
                \edef\yb{\the\pgf@y}%
                \pgfmathtruncatemacro\xbeg{ceil(\xa/\xs)}
                \pgfmathtruncatemacro\xend{floor(\xb/\xs)}
                \if@showgrid@below
                    \foreach \X in {\xbeg,...,\xend} {
                        \node [below,show grid/labels,show grid/xlabels] at (\X,\ya) {\X};
                    }
                \fi
                \if@showgrid@above
                    \foreach \X in {\xbeg,...,\xend} {
                        \node [above,show grid/labels,show grid/xlabels] at (\X,\yb) {\X};
                    }
                \fi
                \pgfmathtruncatemacro\ybeg{ceil(\ya/\ys)}
                \pgfmathtruncatemacro\yend{floor(\yb/\ys)}
                \if@showgrid@left
                    \foreach \Y in {\ybeg,...,\yend} {
                        \node [left,show grid/labels,show grid/ylabels] at (\xa,\Y) {\Y};
                    }
                \fi
                \if@showgrid@right
                    \foreach \Y in {\ybeg,...,\yend} {
                        \node [right,show grid/labels,show grid/ylabels] at (\xb,\Y) {\Y};
                    }
                \fi
            \end{pgfonlayer}
        \fi
    \end{scope}
}
\makeatother
\tikzset{every show grid/.style={show grid/keep bb}%
}%

\graphicspath{{./figures/images/}}
\begin{document}
\title{POAM: Probabilistic Online Attentive Mapping\\for Efficient Robotic Information Gathering}
\author{
\authorblockN{Weizhe Chen, Lantao Liu and Roni Khardon}
\authorblockA{Luddy School of Informatics, Computing, and Engineering\\
Indiana University, Bloomington, IN 47408, USA \\
Emails: chenweiz@iu.edu, lantao@iu.edu, rkhardon@iu.edu}
}
\maketitle
\begin{abstract}
Gaussian Process (GP) models are widely used for Robotic Information Gathering (RIG) in exploring unknown environments due to their ability to model complex phenomena with non-parametric flexibility and accurately quantify prediction uncertainty. Previous work has developed informative planners and adaptive GP models to enhance the \emph{data efficiency} of RIG by improving the robot's sampling strategy to focus on informative regions in non-stationary environments. However, \emph{computational efficiency} becomes a bottleneck when using GP models in large-scale environments with limited computational resources. We propose a framework -- Probabilistic Online Attentive Mapping (POAM) -- that leverages the modeling strengths of the non-stationary Attentive Kernel while achieving \emph{constant-time} computational complexity for online decision-making. POAM guides the optimization process via variational Expectation Maximization, providing constant-time update rules for inducing inputs, variational parameters, and hyperparameters. Extensive experiments in active bathymetric mapping tasks demonstrate that POAM significantly improves computational efficiency, model accuracy, and uncertainty quantification capability compared to existing online sparse GP models.
\end{abstract}
\IEEEpeerreviewmaketitle
\section{Introduction}\label{sec:introduction}
Robotic Information Gathering (RIG) is a fundamental research domain in robotics, focused on how robots can efficiently collect informative data to construct an accurate model of an unknown target function while adhering to robot embodiment constraints. RIG has broad applications, including autonomous environmental monitoring, active 3D reconstruction and inspection, and robotic exploration of uncharted environments~\cite{thrun2002probabilistic, dunbabin2012robots, hollinger2014sampling, schlotfeldt2018anytime, schlotfeldt2021resilient, arora2019multimodal, ghaffari2019sampling, zhu2021online, kantaros2021sampling, fernandez2022informative, xu2023care}.

To achieve effective active information acquisition, certain key features are desired in a model. Firstly, it should efficiently quantify uncertainty in its predictions to guide informed data collection, necessitating a probabilistic approach. Secondly, to facilitate real-time decision-making with limited on-board computational and memory resources, an online model is essential. Finally, the model should exhibit attentiveness in two specific ways: by directing the robot to focus on areas with high modeling errors, and by prioritizing the modeling of the most characteristic aspects of the target function, given the limited computational resources. Considering these requirements, we conclude that a \emph{probabilistic}, \emph{online}, and \emph{attentive} model is necessary.

In spatial modeling, Gaussian Processes (GPs) are widely adopted due to their non-parametric flexibility and precise uncertainty quantification capabilities~\cite{rasmussen2005mit, flaspohler2019information, fernandez2022informative, booth2023informative, kontoudis2023adaptive, chen2023adaptive}. While previous research has primarily focused on developing informative planners and non-stationary kernels to enhance the \emph{data efficiency} of RIG, there has been a significant gap in addressing \emph{computation efficiency}~\cite{ma2017informative, jang2020multi, jakkala2023multi}. This oversight presents a bottleneck in large-scale environments, especially when computational resources are limited.

In this work, we aim to address this critical gap by proposing an online and attentive GP model, building upon the recently introduced Attentive Kernel (AK)~\cite{chen2022ak}, which effectively guides robots to focus on informative regions. This paper tackles the computational limitations of standard GP regression, enabling long-term, long-range RIG missions. Our goal is to achieve \emph{constant-time} computational complexity, which is essential for online decision making, without compromising prediction accuracy and precision in uncertainty quantification.

To address the computational challenge, sparse Gaussian Process (GP) methods are widely used to scale GP models for handling large-scale problems effectively~\cite{quinonero2005unifying, hoang2015unifying, bui2017unifying}. While numerous efforts aim to reduce the computational cost of GPs, existing methods often fail when the data or kernel are non-stationary or when data is collected sequentially in a stream, as in RIG. Initially, we attempted to integrate the AK with established online SGP methods, such as Streaming Sparse GPs (SSGP) introduced by \citet{bui2017streaming} and Online Variational Conditioning (OVC) by \citet{maddox2021conditioning}. However, these approaches present several limitations that motivated us to develop a novel framework. Specifically, when these methods are used with the AK they face difficulties in adapting the input-dependent lengthscale, essentially degenerating to a stationary kernel. SSGP also has suboptimal runtime performance due to the complexity of its online objective function calculation. Additionally, the numerical stability of this method was occasionally compromised, leading to Cholesky decomposition errors. With OVC, the critical aspect of hyperparameter optimization is absent from their discussion, posing a gap in understanding and potentially impacting the overall efficacy of the method.

To overcome these limitations, we introduce the Probabilistic, Online, and Attentive Mapping~(POAM) framework for learning SGP models with the AK. Our analysis reveals that the primary cause of the existing methods' inability to properly learn the lengthscale is the interplay between the optimization of inducing inputs and the kernel lengthscale within SGP methods. To address this, we propose a solution that strategically guides the optimization process by focusing solely on learning the lengthscale and directly computing the inducing-point locations using a lengthscale-dependent strategy. Furthermore, we recognize that optimizing inducing inputs, variational parameters, and hyperparameters should be orchestrated within a variational Expectation-Maximization~(EM) framework. For efficient online decision-making, we introduce a constant-time update rule for variational parameters, leveraging the additive property of the data-dependent variables in the posterior distribution. Additionally, for hyperparameter optimization, our approach employs mini-batch stochastic optimization on the entire dataset, thereby bypassing the extra computational cost associated with computing the Kullback-Leibler (KL) divergence in an online objective function.

In summary, the contributions of this work are threefold:
\begin{itemize}
  \item We diagnose the failure of SGP methods with non-stationary kernels like AK, providing insights for learning input-dependent lengthscales.
  \item We develop constant-time update rules for inducing inputs, variational parameters, and hyperparameters in the SGP framework, enabling real-time decision-making.
  \item We show that the proposed POAM framework outperforms existing methods in modeling accuracy, uncertainty quantification, and computational efficiency in active bathymetric mapping tasks. We also open-source the code for future research\footnote{\url{https://github.com/Weizhe-Chen/POAM}}.
\end{itemize}

\section{Related Work }%
\label{sec:literature}
\topic{Robotic Information Gathering}
Mobile robots have been widely utilized as autonomous data-collecting devices, significantly advancing scientific research, especially in remote or hazardous  environments~\citep{khatib2016ocean, dunbabin2012robots, li2020exploration}. These robots find applications in diverse domains, including environmental mapping and monitoring~\citep{hollinger2013active,girdhar2014autonomous,hitz2017adaptive,manjanna2018heterogeneous,sung2019competitive}, search and rescue missions~\cite{meera2019obstacle,moon2022tigris,niroui2019deep}, and 3D reconstruction projects~\citep{kompis2021informed,zhu2021online,zeng2023efficient}.

In RIG research, the primary focus is on informative planning -- strategically planning action sequences or learning policies to acquire valuable data~\citep{popovic2024robotic}. This emphasis has led to various planners, including recursive greedy approaches~\citep{singh2007efficient,meliou2007nonmyopic,binney2013optimizing}, dynamic programming methods~\citep{low2009information,cao2013multi}, mixed-integer quadratic programming solutions~\citep{yu2014correlated}, sampling-based algorithms~\citep{hollinger2014sampling,ghaffari2019sampling,schmid2020efficient,kantaros2021sampling,schlotfeldt2018anytime,arora2019multimodal,best2019dec,morere2017sequential}, and trajectory optimization techniques~\citep{marchant2012bayesian,bai2016information,di2021multi}. Additionally, efforts have been made to enhance the computational efficiency of information-theoretic objective functions in RIG~\citep{charrow2015information,zhang2020fsmi,carrillo2015monotonicity,xu2023care}. Last but not least, there is a small but growing body of work focusing on the development of probabilistic models and their multi-robot extensions for RIG~\citep{ouyang2014multi,ma2017informative,luo2018adaptive,jang2020multi,jin2022adaptive,jakkala2023multi}. An independent yet closely related research direction is \emph{ergodic search}, which plans a continuous sampling trajectory such that the time a robot spends in a region is generally proportional to the information available in that region~\cite{sun2024fast,dong2023time,ren2023pareto}.

\topic{Online Sparse Gaussian Processes}
Sparse Gaussian process reduce the computational complexity of GP by performing inference through a reduce set of data points or pseudo data points
\cite{quinonero2005unifying,titsias2009variational,hensman2013gaussian,sheth2015sparse}.
For the online setting 
\citet{csato2002sparse} employed Expectation Propagation~(EP) for inference and used a projection method to achieve sparsity, but they did not estimate hyperparameters. In contrast, \citet{bui2017streaming} introduced Streaming Sparse Gaussian Processes (SSGP), which are capable of estimating hyperparameters through an online evidence lower bound~(ELBO). They observed that EP performed worse than variational inference~(VI). A different perspective is provided by \citet{maddox2021conditioning}, who interpreted SSGP as Sparse Gaussian Process Regression (SGPR) on an augmented dataset. This interpretation simplifies derivations and implementation but does not address hyperparameter optimization. Additionally, \citet{stanton2021kernel} enhanced the computational efficiency of SGPR by adopting a structured covariance function, which may limit the model's flexibility.

\topic{Relationship Between Existing Work and Our Work}

Our work is closely related to SSGP~\cite{bui2017streaming} and OVC~\cite{maddox2021conditioning}. Compared to SSGP, 
in addition to the difference in optimization objective,
our method has three key differences. First, we use a different update rule for variational parameters, leading to improved performance and numerical stability. Second, while SSGP initializes inducing inputs by randomly sampling from old inducing inputs and new training inputs and then optimizes the inducing point locations using the online ELBO, we employ pivoted Cholesky decomposition for inducing point selection without further optimization. This approach is faster and avoids training difficulties. Third, for hyperparameter optimization, SSGP optimizes the online ELBO on new batches until convergence and discards the data. In contrast, we sample mini-batches from the entire dataset and use stochastic optimization for a fixed number of iterations, leveraging the fact that RIG is not a streaming problem and all data is available for training. Compared to OVC, our method has two differences: it saves different data-dependent terms for updating variational parameters, resulting in significantly better performance in the experiments, and it addresses hyperparameter optimization, a crucial aspect for RIG that OVC does not discuss extensively.

\section{Background}%
\label{sec:background}

\subsection{Problem Formulation}%
\label{sub:problem_formulation}
We consider a regression problem where the robot observes training data $\mathbb{D}=\{(\mathbf{x}_{i},y_{i})\}_{i=1}^{N}$. Here, $\mathbf{x}_{i}\in\real^{D}$ is the $i$-th input and $y_{i}\in\real$ is the corresponding target value. We build a GP model using the training dataset to predict the outputs $y_{\star}$ given any test input $\mathbf{x}_{\star}$, while taking into account the uncertainty about the prediction for active information gathering. The number of training data $N$ grows as the robot collects more data. Our goal is to develop an algorithm that can efficiently update the GP model and make predictions in \emph{constant} time for online decision making.

\subsection{Attentive Kernel~(AK)}%
\label{sub:attentive_kernel}
We use the Attentive Kernel~(AK)~\cite{chen2022ak}, which is \emph{non-stationary} and has been shown to provide improved prediction accuracy and uncertainty quantification compared to commonly used stationary kernels. The kernel is defined as:
\begin{equation*}
k(\mathbf{x},\mathbf{x}')=\alpha\bar{\mathbf{w}}^{\T}\bar{\mathbf{w}}'\sum_{m=1}^{M}\bar{w}_{m}\bar{w}_{m}'k_{m}(\mathbf{x},\mathbf{x}'),
\end{equation*}
where $\alpha$ is the kernel amplitude, $\bar{\mathbf{w}} = \nicefrac{\mathbf{w}_{\bm{\theta}}(\mathbf{x})}{\norm{\mathbf{w}_{\bm{\theta}}(\mathbf{x})}}$ is the normalized weight vector, and $\bar{w}_{m}$ is its $m$-th element. The vector-valued function $\mathbf{w}_{\bm{\theta}}(\mathbf{x}): \mathbb{R}^{D} \mapsto [0,1]^{M}$ is the output of a neural network parameterized by $\bm{\theta}$.

We use Gaussian kernels with fixed lengthscales $\ell_{m}$ linearly spaced from $\ell_{\text{min}}$ to $\ell_{\text{max}}$ as the base kernels:
\begin{equation*}
k_{m}(\mathbf{x},\mathbf{x}') = \exp \left(-\frac{\norm{\mathbf{x} - \mathbf{x}'}^{2}}{2{\ell_{m}}^{2}} \right).
\end{equation*}

AK captures mild spatially-varying variability by adaptively blending multiple base kernels with different length-scales via input-dependent weights $\bar{w}_{m} \bar{w}_{m}'$. It also handles sharp transitions by zeroing out the correlation between two points through the dot product of their membership vectors $\bar{\mathbf{w}}^{\T} \bar{\mathbf{w}}'$. The parameters of the weighting function $\mathbf{w}_{\bm{\theta}}$ are learned from data by maximizing the standard marginal likelihood or the evidence lower bound.

Compared to commonly used stationary kernels in RIG, AK can better capture key characteristic patterns of the underlying environment and provide more informative uncertainty estimates to guide the robot's sampling process, making it particularly suitable for our problem.

\subsection{Gaussian Process Regression~(GPR)}%
\label{sub:gpr}
\topic{Model}
In Gaussian process regression~(GPR), the target values $y$ are assumed to be the latent function values $f(\mathbf{x})$ corrupted by an additive Gaussian white noise:
\begin{equation*}
y=f(\mathbf{x}) + \varepsilon, \quad \varepsilon \sim \mathcal{N}(\varepsilon \mid 0, \sigma^{2}).
\end{equation*}
A zero-mean GP \emph{prior} is placed over the latent function:
\begin{equation}
     f(\mathbf{x})\sim\mathcal{GP}(0,k_{\bm{\theta}}(\mathbf{x},\mathbf{x}')),
\end{equation}
where $k_{\bm{\theta}}(\mathbf{x},\mathbf{x}')$ is the covariance function, a.k.a. kernel function, with kernel parameters $\bm{\theta}$.

\topic{Prediction}
The GP prior and the Gaussian likelihood are \emph{conjugate}, which yields an elegant closed-form solution for the predictive distribution of the latent function $f_{\star}$ at any test input $\mathbf{x}_{\star}$:
\begin{align}
    p(f_{\star}\mid\mathbf{y})&=\mathcal{N}(f_{\star}\mid\mu,\nu),\label{eq:gpr_prediction}\\
    \mu&=\mathbf{k}_{f\star}^{\T}\mathbf{K}_{yy}^{-1}\mathbf{y},\notag\\
    \nu&=k_{\star\star}-\mathbf{k}_{f\star}^{\T}\mathbf{K}_{yy}^{-1}\mathbf{k}_{f\star},\notag
\end{align}
where $\mathbf{k}_{f\star}\in\real^{N}$ is the kernel values between training inputs and the test input, $\mathbf{K}_{yy}=\mathbf{K}_{f f} + \bm{\Sigma}$ is the summation of the training covariance matrix $\mathbf{K}_{ff}$ and the diagonal observational noise matrix $\bm{\Sigma}=\sigma^{2}\mathbf{I}$, and $k_{\star\star}\triangleq k(\mathbf{x}_{\star},\mathbf{x}_{\star})$.

\topic{Optimization}
The prediction quality of GPR depends on the settings of the \emph{hyperparameters} $\bm{\phi}\triangleq[\alpha,\sigma,\bm{\theta}]$, which can be optimized via \emph{model selection} by maximizing the \emph{model evidence}, a.k.a. log \emph{marginal likelihood}:
\begin{align}
    &\max_{\bm{\phi}}\ \log{p_{\bm{\phi}}(\mathbf{y})}=\max_{\bm{\phi}}\ \log \mathcal{N}(\mathbf{y} \mid \mathbf{0}, \mathbf{K}_{yy})\label{eq:gpr_evidence}\\
    =&\max_{\bm{\phi}}\ -\frac{1}{2}\mathbf{y}^{\T}\mathbf{K}_{yy}^{-1}\mathbf{y}-\frac{1}{2}\log|\mathbf{K}_{yy}|-\frac{N}{2}\log(2\pi),\notag
\end{align}
where $|\mathbf{K}_{yy}|$ is the matrix determinant of $\mathbf{K}_{yy}$.
We refer to hyperparameter optimization as \emph{training} hereafter.

\topic{Complexity}
Training complexity of GPR is $\mathcal{O}(N^{3})$ due to the inversion and determinant computation of $\mathbf{K}_{yy}$ and prediction complexity is $\mathcal{O}(N^{2})$ per test input due to matrix multiplications.

\subsection{Sparse Gaussian Process Regression~(SGPR)}%
\label{sub:sgpr}
\topic{Model}
Sparse Gaussian process regression~(SGPR) alleviates the computational burden by approximating the full GP model with a sparse model that shifts the expensive computations to a small set of \emph{inducing points}~\cite{quinonero2005unifying,titsias2009variational,hensman2013gaussian,sheth2015sparse}.
Specifically, the model is augmented by a small number of $M$ \emph{inducing points} $\{(\mathbf{z}_{m}, u_{m})\}_{m=1}^{M}$, where \emph{inducing outputs} $\mathbf{u}=[f(\mathbf{z}_{1}),\dots,f(\mathbf{z}_{M})]^{\T}$ are evaluated at corresponding \emph{inducing inputs} $\mathbf{Z} = [\mathbf{z}_{1},\dots,\mathbf{z}_{M}]^{\T}$.
The augmented model $p(\mathbf{y}, \mathbf{f}, \mathbf{u})$ can be factorized as
\begin{equation*}
  p(\mathbf{y}, \mathbf{f}, \mathbf{u}) = p(\mathbf{y} \mid \mathbf{f})\ p(\mathbf{f} \mid \mathbf{u})\ p(\mathbf{u}).
\end{equation*}
Assuming that the inducing outputs $\mathbf{u}$ effectively capture the information from the training outputs $\mathbf{f}$ (i.e., any other function value and $\mathbf{f}$ are \emph{independent} given $\mathbf{u}$), the posterior predictive distribution of $f_{\star}$ can be expressed as:
\begin{align*}
  p(f_{\star} \mid \mathbf{y}) =& \int p(f_{\star} \mid \mathbf{u}, \mathbf{f})\ {\color{gray}p(\mathbf{f} \mid \mathbf{u}, \mathbf{y}})\ p(\mathbf{u} \mid \mathbf{y})\ {\color{gray}\mathrm{d} \mathbf{f}}\ \mathrm{d} \mathbf{u}\\
  =& \int p(f_{\star} \mid \mathbf{u})\ p(\mathbf{u} \mid \mathbf{y})\ \mathrm{d} \mathbf{u},
\end{align*}
which indicates that prediction can be made by only considering the inducing points.

In practice, however, the assumption that \(\mathbf{u}\) serves as sufficient statistics is unlikely to hold, and the posterior distribution becomes intractable in non-conjugate cases. To address these issues, the posterior distribution is typically approximated by a structured variational distribution:
\begin{equation*}
  p(\mathbf{f}, \mathbf{u} \mid \mathbf{y}) \approx q(\mathbf{f}, \mathbf{u}) = p(\mathbf{f} \mid \mathbf{u})\ q(\mathbf{u}),
\end{equation*}
where \(p(\mathbf{f} \mid \mathbf{u})\) is the conditional prior and \( q(\mathbf{u}) = \mathcal{N}(\mathbf{u} \mid \mathbf{m}, \mathbf{S}) \) is the approximate posterior of the inducing variables, assumed to follow a Gaussian distribution. The parameters \(\mathbf{m}\) and \(\mathbf{S}\) can be optimized via variational inference~\citep{titsias2009variational,hensman2013gaussian,sheth2015sparse}.

\topic{Prediction}
Choosing a Gaussian distribution for $q(\mathbf{u})$ leads to a closed-form solution for the predictive distribution $p(f_{\star} \mid \mathbf{y}) \approx q(f_{\star})$:
\begin{align}
  q(f_{\star}) &= \mathcal{N}\left(f_{\star} \mid \tilde{\mu}, \tilde{\nu}\right),\label{eq:sgpr_prediction}\\
  \tilde{\mu} &= \mathbf{k}_{u \star}^{\T} \mathbf{K}_{u u}^{-1} \mathbf{m},\notag\\
  \tilde{\nu} &= k_{\star \star}-\mathbf{k}_{u \star}^{\T} \mathbf{K}_{u u}^{-1} \mathbf{k}_{u \star}+\mathbf{k}_{u \star}^{\T} \mathbf{K}_{u u}^{-1} \mathbf{S} \mathbf{K}_{u u}^{-1} \mathbf{k}_{u \star}.\notag
\end{align}

\topic{Optimization}
Variational inference optimizes the parameters \(\mathbf{Z}\), \(\mathbf{m}\), and \(\mathbf{S}\) by minimizing the Kullback-Leibler (KL) divergence, \(\mathtt{KL}[q(\mathbf{f}, \mathbf{u}) \| p(\mathbf{f}, \mathbf{u}\,|\,\mathbf{y})]\). This is equivalent to maximizing the evidence lower bound (ELBO)~\cite{titsias2009variational}:
\begin{align}
  \log p(\mathbf{y}) \geq& \int q(\mathbf{f}, \mathbf{u})\ \log \frac{p(\mathbf{y}, \mathbf{f}, \mathbf{u})}{q(\mathbf{f}, \mathbf{u})}\ \mathrm{d} \mathbf{f}\ \mathrm{d} \mathbf{u} \triangleq \mathtt{ELBO},\label{eq:elbo}\\
  \mathtt{ELBO} =& \log \mathcal{N}(\mathbf{y} \mid \mathbf{0}, \mathbf{Q}_{yy})-\frac{1}{2\sigma^{2}}\mathrm{tr}(\mathbf{K}_{ff}-\mathbf{Q}_{ff})\label{eq:collapsed_elbo}.
\end{align}
Here, \(\mathbf{Q}_{ff} = \mathbf{K}_{uf}^{\T} \mathbf{K}_{uu}^{-1} \mathbf{K}_{uf}\) and \(\mathbf{Q}_{yy} = \mathbf{Q}_{ff} + \bm{\Sigma}\). This bound is referred to as the \emph{collapsed} form of the ELBO, as the variational parameters \(\mathbf{m}\) and \(\mathbf{S}\) are analytically marginalized out due to the Gaussian likelihood used in the regression case.

The optimal variational parameters correspond to the collapsed ELBO are given by
\begin{align}
  \mathbf{m} =& \frac{1}{\sigma^{2}}\mathbf{K}_{u u} \mathbf{A}^{-1} \mathbf{K}_{u f} \mathbf{y},\label{eq:sgpr_m}\\
  \mathbf{S} =& \mathbf{K}_{u u} \mathbf{A}^{-1} \mathbf{K}_{u u},\text{ where} \label{eq:sgpr_S}\\
  \mathbf{A} =& \mathbf{K}_{u u} + \frac{1}{\sigma^{2}}\mathbf{K}_{u f}\mathbf{K}_{u f}^{\T}.
\end{align}
Note that, although \(\mathbf{m}\) and \(\mathbf{S}\) are analytically marginalized out and thus do not require optimization, we still need to optimize the inducing inputs \(\mathbf{Z}\) and the hyperparameters \(\bm{\phi}\). This can be achieved simultaneously by maximizing the ELBO.

\topic{Complexity}
The matrix inversion operations in SGPR are $\mathcal{O}(M^{3})$ instead of $\mathcal{O}(N^{3})$ in GPR, which is no longer the bottleneck.
The time complexity of SGPR is $\mathcal{O}(NM^{2})$ due to the matrix multiplications, which is much lower than that of GPR when $M \ll N$.

\subsection{Stochastic Variational Gaussian Processes~(SVGP)}%
\label{sub:svgp}
\topic{Optimization}
The linear complexity of SGPR can be further reduced to constant time using Stochastic Variational Gaussian Processes~(SVGP)~\cite{hensman2013gaussian}.
The ELBO in \Cref{eq:elbo} before marginalizing out the variational parameters can be written in an \emph{uncollapsed} form:
\begin{align}
  \mathtt{ELBO} = \sum_{n=1}^{N}\mathtt{E}_{q(f_{n})}\left[\log p(y_{n} \mid f_{n})\right] - \mathtt{KL}[q(\mathbf{u})\ \|\ p(\mathbf{u})],\label{eq:svgp_elbo}
\end{align}
where the marginal predictive distribution $q(f_{n})$ is given by \Cref{eq:sgpr_prediction}.
The expectation can be computed analytically when the likelihood is Gaussian or approximated by Monte Carlo sampling otherwise~\cite{hoffman2013stochastic}.
Writing the ELBO as a sum of $N$ terms allows us to use mini-batch stochastic optimization to update the parameters.
Specifically, the gradient of the expected log-likelihood term can be approximated by $B$ training data sampled uniformly at random:
\begin{align}
  \frac{N}{B}\sum_{b=1}^{B}\nabla\mathtt{E}_{q(f_{b})}[\log p(y_{b} \mid f_{b})].\label{eq:minibatch}
\end{align}
All the parameters, including the variational parameters $\mathbf{m}$ and $\mathbf{S}$, the inducing inputs $\mathbf{Z}$, and the hyperparameters $\bm{\phi}$, can be optimized by maximizing the uncollapsed ELBO in \Cref{eq:svgp_elbo} using stochastic gradient ascent.

\topic{Prediction}
The resulting variational parameters \(\mathbf{m}\) and \(\mathbf{S}\) can be directly plugged into \Cref{eq:sgpr_prediction} to make predictions, bypassing the need to compute them analytically from the training data, which would take linear time.

\topic{Complexity}
The time complexity becomes \(\mathcal{O}(BM^{2} + M^{3})\) per training iteration and \(\mathcal{O}(M^{3})\) when making predictions on a test input, which is independent of the number of training data \(N\).

\begin{figure}[tb]
  \centering
  \subfloat[Environment]{\includegraphics[width=0.45\linewidth,height=0.3\linewidth,trim={100 30 130 60},clip]{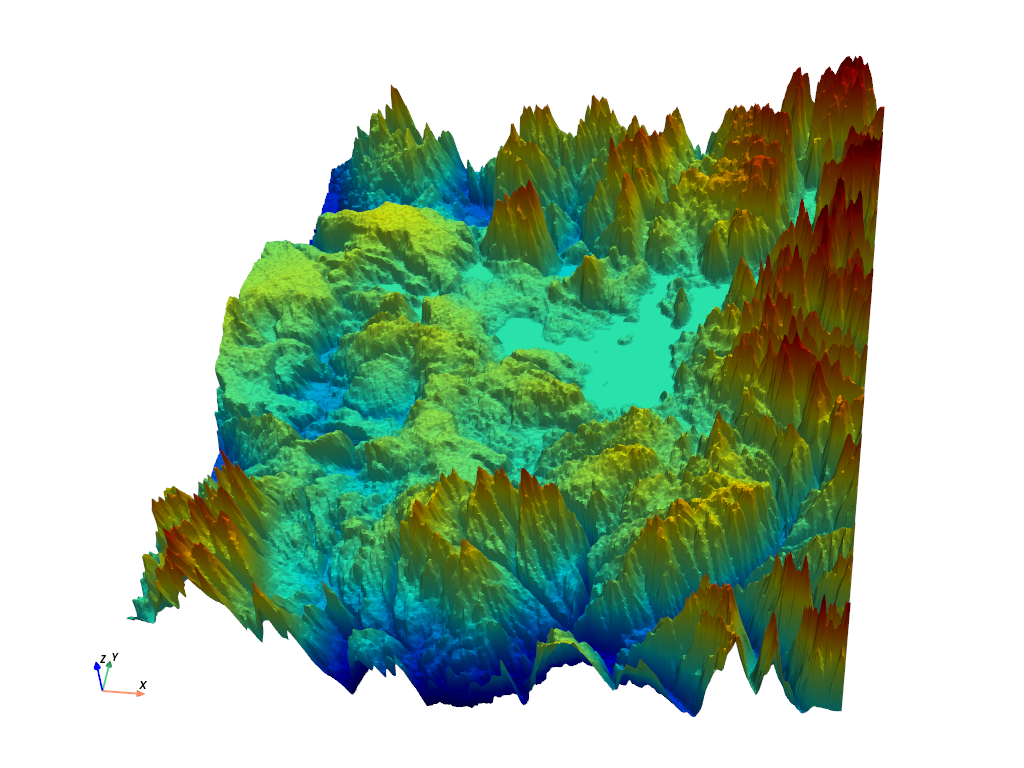}}%
  \subfloat[Dataset]{\includegraphics[width=0.45\linewidth,height=0.3\linewidth,trim={100 30 130 60},clip]{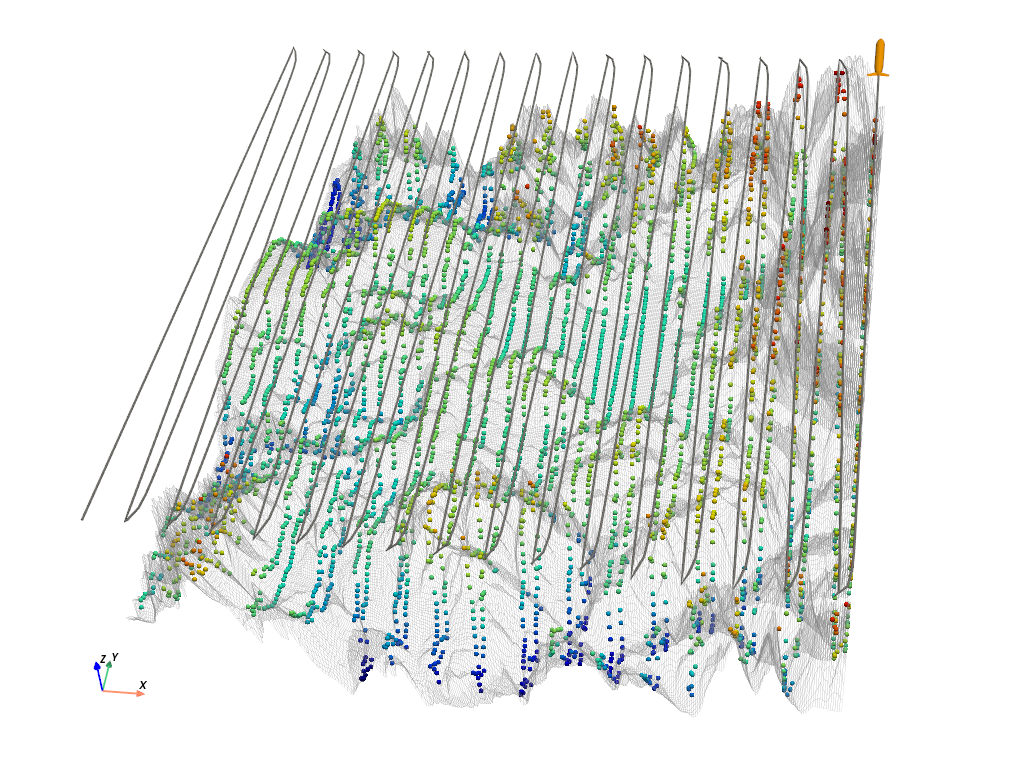}}%
  \caption{\textbf{An elevation dataset sampled via lawnmower path in the environment for illustrative purposes.} (a) The ground-truth elevation map of the environment. Red and blue colors represent high and low elevations, respectively. (b) A dense dataset sampled by lawnmower path in the environment, which is used to train the GP model for elevation mapping.}\label{fig:demo_data}%
\end{figure}

\section{Methodology}%
\label{sec:methodology}
Before introducing the complete Probabilistic, \emph{Online}, and Attentive Mapping~(POAM) framework, we first explain how to achieve probabilistic attentive mapping using SVGP with AK on the \emph{entire} training dataset. Understanding this offline learning case is crucial for developing the online model. For illustrative purposes, we use the environment and dataset shown in \Cref{fig:demo_data} as a running example. The goal in this example is to make predictions that closely match the ground-truth environment using the collected data and to learn the underlying lengthscale map. Specifically, the left and center parts should have a large lengthscale, while the other three sides should have a small lengthscale. We then introduce the online update rules for the inducing inputs, variational parameters, and hyperparameters, enabling the model to be updated incrementally and efficiently as new data arrives.

\subsection{Probabilistic Attentive Mapping with SVGP}%
\begin{figure}[tb]
  \centering
  \subfloat[SVGP Mean\label{fig:smooth_svgp}]{\includegraphics[width=0.45\linewidth,height=0.3\linewidth,trim={80 25 80 80},clip]{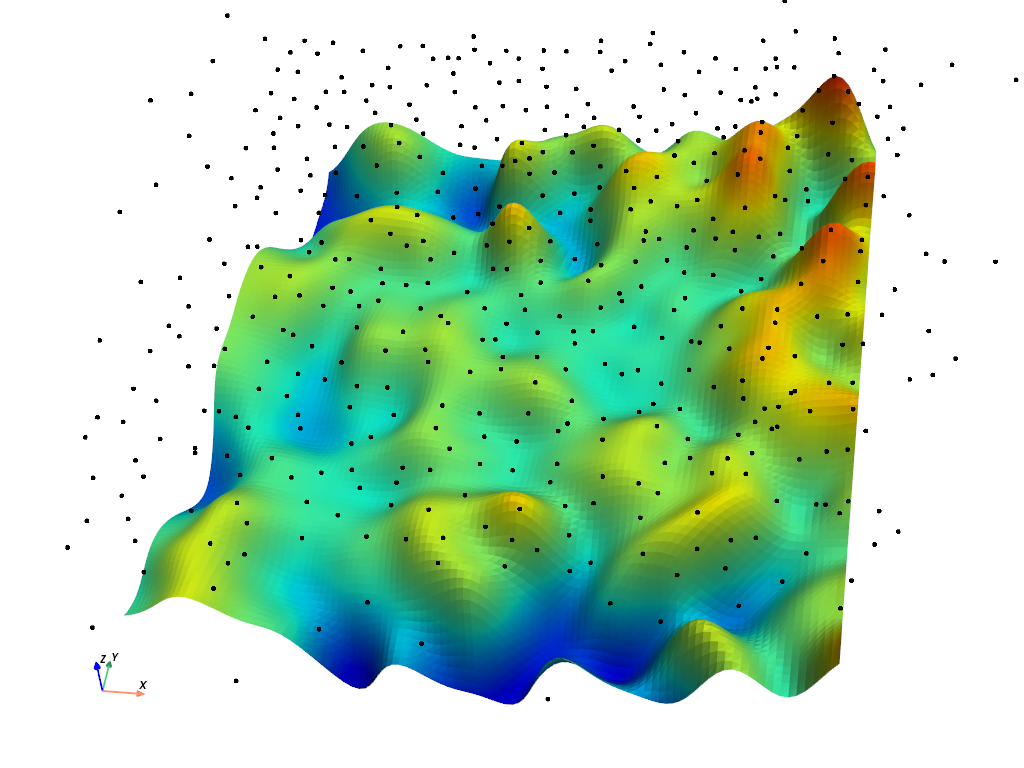}}%
  \subfloat[SVGP Lengthscale\label{fig:flat_ak}]{\includegraphics[width=0.45\linewidth,height=0.3\linewidth,trim={80 25 80 80},clip]{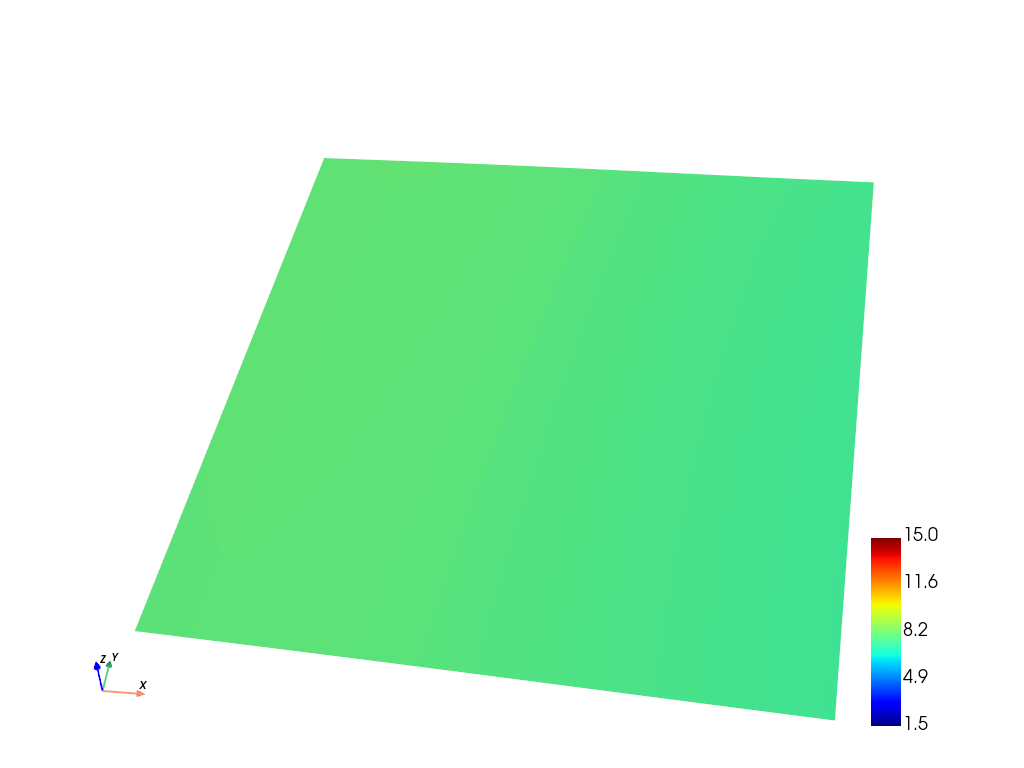}}%
\caption{\textbf{Prediction and lengthscale maps from jointly training all parameters with an Adam optimizer}. (a) The predictive mean map exhibits  uniform smoothness due to the uniform scattering of inducing points (black dots) across the space and the failure to learn an input-dependent lengthscale. (b) The lengthscale map is flat, indicating that the attentive kernel has degenerated to a stationary kernel.}\label{fig:demo_svgp_results}%
\end{figure}

\label{sub:pam}
\topic{Direct Use of SVGP with AK}
A straightforward solution to achieve probabilistic, online, and attentive mapping is to use an AK in SVGP and optimize all parameters using mini-batch stochastic optimization on an \emph{ever-growing} dataset. Thus, directly training an SVGP with AK on the entire dataset is our initial attempt to achieving probabilistic attentive mapping. However, as shown in \Cref{fig:demo_svgp_results}, this method encounters difficulties in learning the appropriate lengthscale map. The primary issue lies in the coupling of inducing input and kernel lengthscale optimization.

During the early stages of training, 
the hyperparameters of the AK have not yet been optimized and the input-dependent lengthscale behaves like a stationary kernel. 
At this stage, optimizing the ELBO in \Cref{eq:svgp_elbo} results in inducing inputs becoming uniformly distributed across the input space, encompassing the training data. When the optimization attempts to allocate a small lengthscale to a region with high variability, it requires a rapid relocation of more inducing inputs to that region. Failure to achieve this relocation incurs a high training loss because regions with small lengthscales require dense inducing-point support. However, gradient-based optimization restricts the movement of inducing inputs in each iteration, making it difficult to meet this requirement promptly. As a result, the optimizer tends to maintain a large lengthscale in AK even in regions of high variability. This ultimately leads to an inability to learn the input-dependent lengthscale, causing the model to degenerate into a stationary kernel, with inducing point locations becoming uninformative.

\topic{Pivoted Cholesky Decomposition}
Building upon the aforementioned observations, we introduce a novel strategy to guide the optimization process: \emph{focusing solely on learning the lengthscale while directly computing the inducing input locations}. 
For the latter, we leverage Pivoted Cholesky Decomposition (PCD)~\cite{harbrecht2012low}, as proposed by \citet{burt2020convergence}
and used in OVC~\cite{maddox2021conditioning},
to efficiently select inducing points from the training data.

Specifically, PCD computes a low-rank Cholesky factorization of a positive-definite matrix by iteratively selecting pivots to compute the permutation matrix for rearranging rows and columns. The algorithm starts with a residual matrix and a permutation matrix, which are initialized as the input matrix and an identity matrix, respectively. Iteratively, the algorithm selects pivots (the maximum diagonal elements of the residual matrix) to construct the Cholesky factor and update the residual matrix. This process continues until a specified rank is achieved or the remaining diagonal elements fall below a predefined error tolerance. Dynamic reordering of rows and columns using a permutation matrix derived from the selected pivot occurs throughout the iterations. With a computational complexity of $\mathcal{O}(NR^{2})$, where $R$ is the rank of the decomposition, PCD strikes a balance between numerical accuracy and efficiency.

When applied to inducing input selection, the input is the kernel matrix $\mathbf{K}_{f f}$, and the output is the low-rank factor of $\mathbf{K}_{f f} \approx \mathbf{L} \mathbf{L}^{\T}$ along with the selected pivots. These pivots are then used to index the training data, enabling the selection of corresponding training inputs as the inducing inputs.

Intuitively, PCD allocates more inducing inputs in complex regions (small-lengthscale regions) and fewer inducing inputs in simple regions (large-lengthscale regions). This results in a more attentive inducing input distribution, enhancing the model's adaptability to spatially-varying complexities. Theory-wise, \citet{burt2020convergence} derived upper bounds on the KL-divergence between the approximate posterior and the true posterior, which depend on either the \emph{trace} or the \emph{largest eigenvalues} of the low-rank Nystr\"om approximation error. PCD provides a simple way to compute a low-rank approximation to a positive-definite matrix such that the trace error is rigorously controlled. It also enjoys exponential convergence rates when the eigenvalues of the full matrix exhibit a fast exponential decay~\cite{harbrecht2012low}. These theoretical properties, along with the computational efficiency and numerical stability, make PCD a good choice for inducing input updates.

\topic{Analytic Computation of the Variational Parameters}
As shown in the ablation study in \Cref{sub:ablation_study}, using PCD for inducing input selection and jointly optimizing all other parameters still results in poor performance. This is because the variational parameters $\mathbf{m}$ and $\mathbf{S}$ depend on the inducing inputs. When inducing inputs are updated instantaneously rather than being slowly optimized by gradient descent, the gradient optimization of the variational parameters cannot adjust accordingly, leading to poor performance. To resolve this, we also compute the variational parameters analytically using the closed-form expressions in \Cref{eq:sgpr_m,eq:sgpr_S}.

\begin{figure}[tb]
  \centering
  \subfloat[PAM Mean\label{fig:pam_mean}]{\includegraphics[width=0.45\linewidth,height=0.3\linewidth,trim={80 30 130 20},clip]{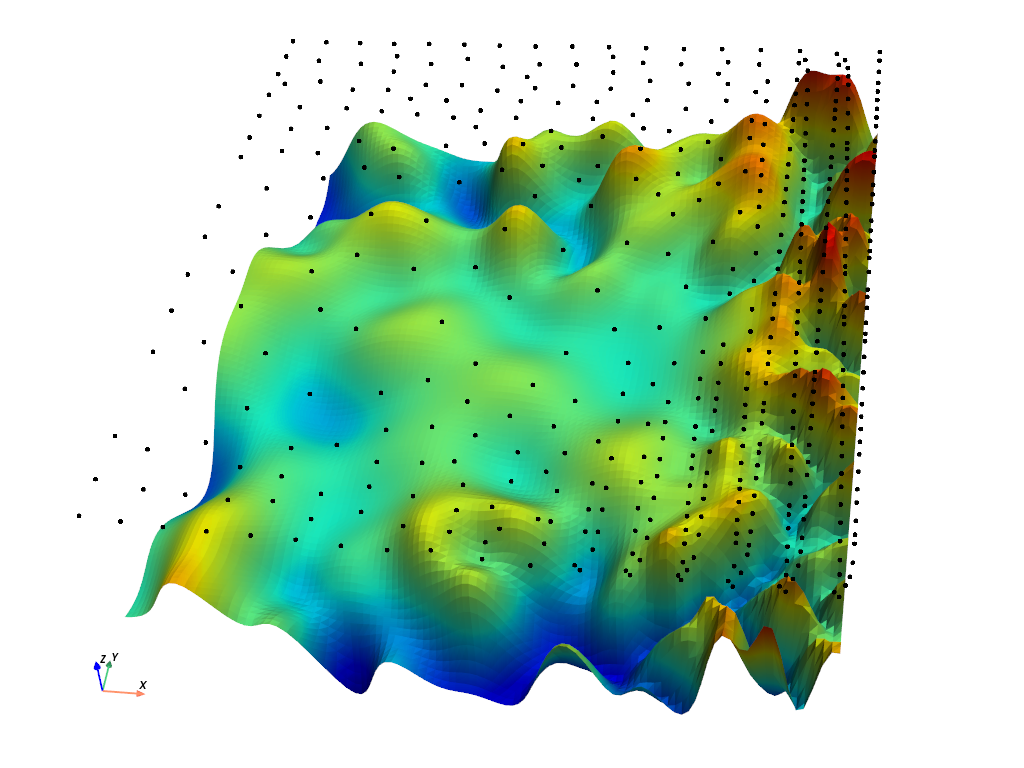}}%
  \subfloat[PAM Lengthscale\label{fig:pam_lenscale}]{\includegraphics[width=0.45\linewidth,height=0.3\linewidth,trim={80 30 80 100},clip]{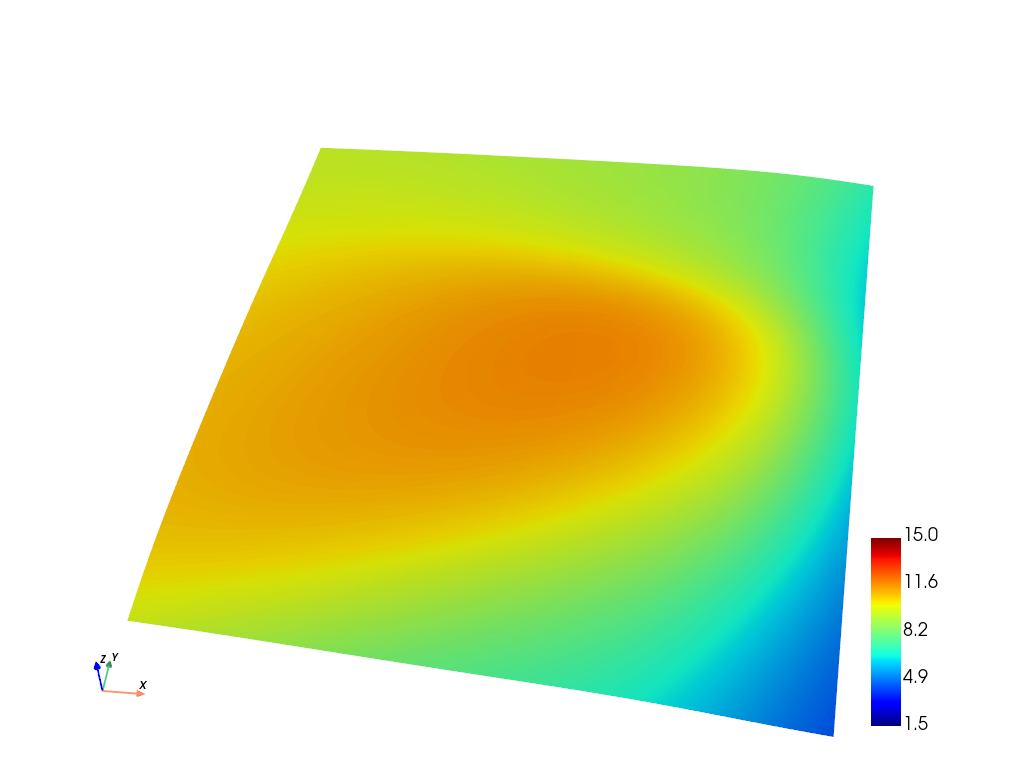}}%
  \caption{\textbf{Prediction and lengthscale maps from the proposed Probabilistic Attentive Mapping~(PAM) training paradigm}. (a) A higher density of inducing points is allocated to the complex region, enabling the predictive mean to capture finer elevation details. (b) The learned lengthscale map delineates the relatively smooth area on the left (large lengthscale) from the highly varying region near the right boundary (small lengthscale).}\label{fig:demo_pam_results}%
\end{figure}

\topic{Variational Expectation Maximization Training}
To account for the interdependence of inducing inputs, variational parameters, and hyperparameters, we adopt a variational Expectation-Maximization (EM) approach to update these three sets of parameters. In the E-step, we fix the hyperparameters and sequentially update the inducing inputs and variational parameters. In the M-step, we fix the inducing inputs and variational parameters while optimizing the hyperparameters. This systematic updating process, when applied to SVGP with AK, is referred to as {\em Probabilistic Attentive Mapping~(PAM)}.

The effectiveness of PAM is showcased in \Cref{fig:demo_pam_results}, where it demonstrates the ability to learn input-dependent lengthscales and strategically allocate more inducing inputs in high-variability regions, leading to more accurate predictions.

However, while PAM excels in learning input-dependent lengthscales, its suitability for online learning is limited due to the non-incremental nature of variational parameter and inducing input computations. In the subsequent sections, we develop constant-time update rules specifically designed for inducing inputs and variational parameters.

\subsection{Online Update of Inducing Inputs}%
\label{sub:online_update_inducing_inputs}

Suppose we have $M$ existing inducing inputs $\mathbf{Z'}$ selected from old training data $\mathbf{X'}$ and now receive $N_{\text{new}}$ new training data $\mathbf{X} \in \mathbb{R}^{N_{\text{new}} \times D}$. Updating the inducing inputs by concatenating the old and new training data $[\mathbf{X'}^{\T}, \mathbf{X}^{\T}]^{\T}$ and running PCD again on the resulting $N \times N$ kernel matrix requires linear time complexity with respect to the number of training data $N$, which is not ideal for online learning.

To address this limitation, we opt for a more efficient \emph{recursive} update rule for inducing inputs. Specifically, we concatenate the old inducing inputs and new training data $[\mathbf{Z'}^{\T}, \mathbf{X}^{\T}]^{\T}$ and perform PCD on the resulting square kernel matrix of size $M + N_{\text{new}}$, which is independent of the total number of training data. Note that $M + N_{\text{new}} \ll N$, and this recursive update rule allows us to efficiently update the inducing inputs in constant time complexity with respect to $N$, making it ideal for online learning.

\subsection{Online Update of Variational Parameters}%
\label{sub:online_update_variational_parameters}

Recomputing the variational parameters using \Cref{eq:sgpr_m,eq:sgpr_S} after receiving new data involves linear time complexity with respect to the number of training data. For real-time decision-making, we also need an incremental update strategy for the variational parameters. The key idea is to preserve the old data-dependent terms and compute only the new data-dependent terms related to the recently acquired data. 

Denoting old variables with a prime symbol and explicitly writing out the data-dependent terms, we have:
\begin{align}
  \begin{bmatrix}
    \mathbf{K}_{u f'} & \mathbf{K}_{u f}
  \end{bmatrix}
  \begin{bmatrix}
    \mathbf{y'} \\
    \mathbf{y}
  \end{bmatrix}
  &= \underbracket{\mathbf{K}_{u f'} \mathbf{y'}}_{\text{old}} + \underbracket{\mathbf{K}_{u f} \mathbf{y}}_{\text{new}}, \\
  \begin{bmatrix}
    \mathbf{K}_{u f'} & \mathbf{K}_{u f}
  \end{bmatrix}
  \begin{bmatrix}
    \mathbf{K}_{f' u} \\
    \mathbf{K}_{f u}
  \end{bmatrix}
  &= \underbracket{\mathbf{K}_{u f'} \mathbf{K}_{f' u}}_{\text{old}} + \underbracket{\mathbf{K}_{u f} \mathbf{K}_{f u}}_{\text{new}}.
\end{align}
The \emph{additive} property of the data-dependent terms allows us to update the variational parameters incrementally. However, if we save the old data-dependent terms, $\mathbf{K{\color{red}'}}_{{\color{red}u'} f'}\mathbf{y'}$ and $\mathbf{K{\color{red}'}}_{{\color{red}u'} f'} \mathbf{K{\color{red}'}}_{f' {\color{red}u'}}$, they are computed using the old inducing inputs $\mathbf{u'}$ and hyperparameters $\bm{\phi'}$. To address this issue, we use a projection matrix to transform the saved terms to align with the new inducing inputs and hyperparameters. Formally, we would like to find $\mathbf{P}\in\real^{M\times{M}}$ such that
\begin{align}
    \mathbf{P}^{\top} \mathbf{K'}_{u' f'} \mathbf{y'} &= \mathbf{K}_{u f'} \mathbf{y'},\\
    \mathbf{P}^{\top} \mathbf{K'}_{u' f'} \mathbf{K'}_{f' u'} \mathbf{P} &= \mathbf{K}_{u f'} \mathbf{K}_{f' u'}.
\end{align}
This requires solving the following linear system for $\mathbf{P}$:
\begin{align}
  \mathbf{P}^{\top} \mathbf{K'}_{u' f'} = \mathbf{K}_{u f'} \text{, or equivalently, } \mathbf{K'}_{f' u'} \mathbf{P} = \mathbf{K}_{f' u}.
\end{align}
Since $\mathbf{K'}_{f' u'}$ is not a square matrix, the solution can be computed using the pseudo inverse:
\begin{align}
  \mathbf{P} =& (\mathbf{K'}_{u' f'}\mathbf{K'}_{f' u'})^{-1} \mathbf{K'}_{u' f'} \mathbf{K}_{f' u}.\label{eq:projection_matrix}
\end{align}
The computation of $\mathbf{K'}_{f' u'}$ and $\mathbf{K}_{f' u}$ requires all the old inputs $\mathbf{X'}$ before the current time step, which still requires linear time complexity with respect to the number of training data. To bypass this issue, we can choose a small set of ``representative'' old inputs for computing the kernel matrices. One computationally convenient choice is to use the old inducing inputs. By replacing $\mathbf{K'}_{f' u'}$ with $\mathbf{K'}_{u' u'}$ and $\mathbf{K}_{f' u}$ with $\mathbf{K}_{u' u}$ in \Cref{eq:projection_matrix}, the projection matrix $\mathbf{P}$ is given by:
\begin{align}
    \mathbf{P} &= (\mathbf{K'}_{u' u'} \mathbf{K'}_{u' u'})^{-1} \mathbf{K'}_{u' u'} \mathbf{K}_{u' u}, \\
    &= \mathbf{K'}_{u' u'}^{-1} \mathbf{K}_{u' u}.
\end{align}
Here, the first term is the inverse of the self-covariance matrix of the old inducing inputs, and the second term is the cross-covariance matrix between the old inducing inputs and the new inducing inputs.

In summary, we can update the variational parameters incrementally by adding the projected old data-dependent terms to the new data-dependent terms:
\begin{align}
  \mathbf{P}^{\top} \underbracket{\mathbf{K'}_{u' f'} \mathbf{y'}}_{\text{saved}} + \mathbf{K}_{u f} \mathbf{y},\label{eq:poam_update_1}\\
  \mathbf{P}^{\top} \underbracket{\mathbf{K'}_{u' f'} \mathbf{K'}_{f' u'}}_{\text{saved}} \mathbf{P} + \mathbf{K}_{u f} \mathbf{K}_{f u}.\label{eq:poam_update_2}
\end{align}
We then plug these updated terms into \Cref{eq:sgpr_m,eq:sgpr_S} to update the variational parameters.

It is important to note that although \Cref{eq:poam_update_1,eq:poam_update_2} resemble the projection-view of OVC, they actually store distinct sets of data-dependent terms: $\mathbf{K'}_{u' f'} \mathbf{\Sigma}^{-1} \mathbf{y'}$ and $\mathbf{K'}_{u' f'} \mathbf{\Sigma}^{-1} \mathbf{K'}_{f' u'}$. This difference results in significantly better performance of POAM in our experiments.

\subsection{Online Update of Hyperparameters}%
\label{sub:online_update_hyperparameters}

For the online update of hyperparameters, our key insight is that RIG does not conform to a strict streaming problem. Typically, the robot retains access to all past data accumulated and stored onboard. To take advantage of this, we leverage the entire training set for hyperparameter updates while maintaining constant training complexity with respect to the number of training data.

Given that SVGP supports mini-batch stochastic optimization, we incorporate new data by appending it to the continually expanding training set. Hyperparameter optimization is then performed for a specified number of steps using \Cref{eq:svgp_elbo}, with gradients approximated through a Monte-Carlo estimate, as described in \Cref{eq:minibatch}, computed from a mini-batch of randomly selected samples.

While this approach reduces the number of updates on the newly collected data because it may not be sampled in the mini-batch, this turns out to be advantageous in the context of RIG problems. During the initial phases, when the robot lacks sufficiently representative data of an unknown environment, frequent hyperparameter updates can prematurely lead the robot into an exploitative mode. Deliberately slowing down hyperparameter updates, especially for the lengthscale, allows the robot to initially explore the environment and gather more representative data for refining its model.

\subsection{Probabilistic, Online, and Attentive Mapping (POAM)}%
\label{sub:poam}
\begin{algorithm}[tb]
  \caption{\textbf{RIG with POAM}}\label{alg:overview}
  Collect $\mathbf{X}_{0}$ and $\mathbf{y}_{0}$ by following a pilot survey path\;
  Initialize $\mathbf{Z}_{0}$, $\mathbf{m}_{0}$, $\mathbf{S}_{0}$, and $\bm{\phi}_{0}$\;
  Decision epoch $t=0$\;
  \While{not collected enough number of samples}{
    Decision epoch $t=t+1$\;
    Select an informative waypoint\;
    Collect samples $\mathbf{X}_{t}$ and $\mathbf{y}_{t}$ by navigating to the selected informative waypoints.\;
    Concatenate $\mathbf{Z}_{t-1}$ and $\mathbf{X}_{t}$\;
    Update $\mathbf{Z}_{t}$ via PCD on the kernel matrix computed on the concatenated inputs\;
    Update the data-dependent terms using \Cref{eq:poam_update_1,eq:poam_update_2}\;
    Update $\mathbf{m}_{t}$ and $\mathbf{S}_{t}$ via \Cref{eq:sgpr_m,eq:sgpr_S}\;
  Optimize hyperparmeters $\bm{\phi}$ on all data by mini-batch stochastic gradient ascent on \Cref{eq:svgp_elbo}\;
  }
\end{algorithm}

The overarching POAM framework is outlined in \Cref{alg:overview}. Initially, the robot conducts a pilot survey to collect training data, which is essential for computing normalizing statistics for data preprocessing and initializing the GP model. To achieve this, the initial path should cover various locations within the workspace boundaries to gather representative samples. During each decision epoch, the robot selects informative waypoints and collects new samples while navigating to these waypoints. The newly acquired training inputs, along with the previous inducing inputs, are then used to update the inducing inputs via PCD (\Cref{sub:online_update_inducing_inputs}). Once the inducing inputs are updated, the variational parameters are adjusted using the online updates described in \Cref{sub:online_update_variational_parameters}. Finally, hyperparameters are optimized by performing mini-batch stochastic gradient ascent on the ELBO for several steps (\Cref{sub:online_update_hyperparameters}). Overall, POAM enables accurate and efficient probabilistic mapping of the unknown target function.

\section{Experiments}%
\label{sec:experiments}
We evaluate the proposed POAM framework's accuracy, uncertainty quantification capability, and computational efficiency through extensive experiments. In benchmarking, POAM is compared with two state-of-the-art online sparse GP models or their enhanced versions. The results, both quantitative and qualitative, are discussed in \Cref{sub:results}. Additionally, an ablation study is conducted to validate the importance of the POAM components, detailed in \Cref{sub:ablation_study}.

\subsection{Experiment Setup}%
\label{sub:experiment_setup}

\topic{Task Setting}
Consider an Autonomous Underwater Vehicle~(AUV) following a Dubins' car kinematic model on a fixed altitude plane, with a maximum linear velocity of 1 m/s and a control frequency of 10 Hz. The AUV is equipped with a single-beam range sensor that collects noisy elevation measurements at 3 Hz, with unit Gaussian white noise. The probabilistic model's inputs are two-dimensional sampling locations, and it can predict elevation at any query location along with the corresponding prediction uncertainty. In this active bathymetric mapping task, the AUV aims to minimize elevation prediction error efficiently, given a budget of 5000 samples. A superior method should achieve a lower prediction error by the end of the task and show a faster reduction in the prediction error curve. For simplicity, all compared methods use the same planner proposed in \cite{chen2022ak}. This planner evaluates prediction entropy at 2000 random candidate locations and 
selects a point maximizing high-entropy and minimizing distance to the current location of the robot 
as the next informative waypoint.
As discussed in Section~\ref{sub:robustness} the conclusions of the experiments are still valid when the planner is changed. 

\begin{figure}[tb]
  \centering
  \includegraphics[width=0.7\linewidth]{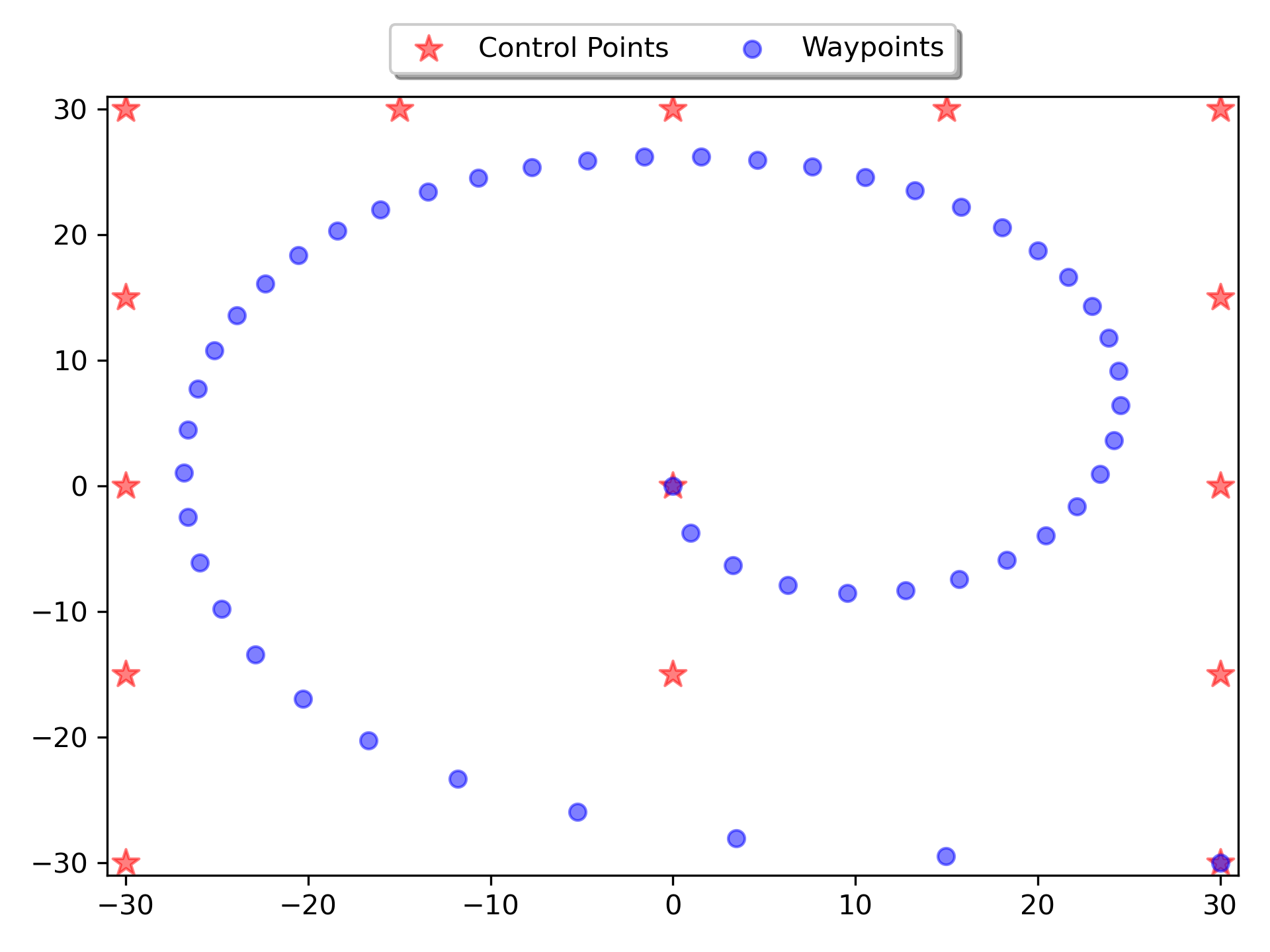}
  \caption{\textbf{Illustration of the initial waypoints generated by a B\'ezier curve}.}\label{fig:pilot_survey}%
\end{figure}

\begin{table*}[tb]
    \centering\small
    \caption{\textbf{Benchmarking Results}. The table shows the average performance across all decision epochs for the proposed POAM and the compared baselines in four different environments. Metrics include SMSE, MSLL, and training time. The best-performing method is highlighted in bold. Results are averaged over 10 runs, with the standard deviation indicated by the plus-minus sign. Superscripts and subscripts denote maximum and minimum values, while arrows indicate the direction of improvement.}\label{tab:quantitative}
    \begin{tabular}{
            rcc
            S[table-format=1.2(3)e2]
            S[table-format=1.2(3)e2]
            S[table-format=1.2(3)e2]
            S[table-format=1.2(3)e1]
            S[table-format=1.2(3)e1]
            S[table-format=1.2(3)e1]
        }
        \toprule
        {\textbf{Name}} & {\textbf{Environment}} & {\textbf{Method}} & {\textbf{Averaged SMSE}$\downarrow^{1}_{0}$} & {\textbf{Averaged MSLL}$\downarrow^{0}$} & {\textbf{Averaged Time (s)}$\downarrow_{0}$}\\
        \midrule
        \multirow{4}{*}{\textbf{Env1}} &
        \multirow{4}{*}{\includegraphics[width=0.1\linewidth,trim={0 0 0 60},clip]{env_n44w111}}
                                       &  SSGP++    &    3.22(0.17)e-01 &    -6.52(0.78)e-01 &    1.57(0.02)e+00\\
                                       && OVC       &    3.18(0.11)e-01 &    -6.16(0.18)e-01 &    1.18(0.00)e+00\\
                                       && OVC++     &    3.13(0.21)e-01 &    -6.70(0.24)e-01 & \B 8.51(0.10)e-01\\
                                       && POAM      & \B 3.09(0.11)e-01 & \B -7.08(0.11)e-01 & \B 8.56(0.09)e-01\\
                                       \midrule
        \multirow{4}{*}{\textbf{Env2}} &
        \multirow{4}{*}{\includegraphics[width=0.1\linewidth,trim={0 0 0 60},clip]{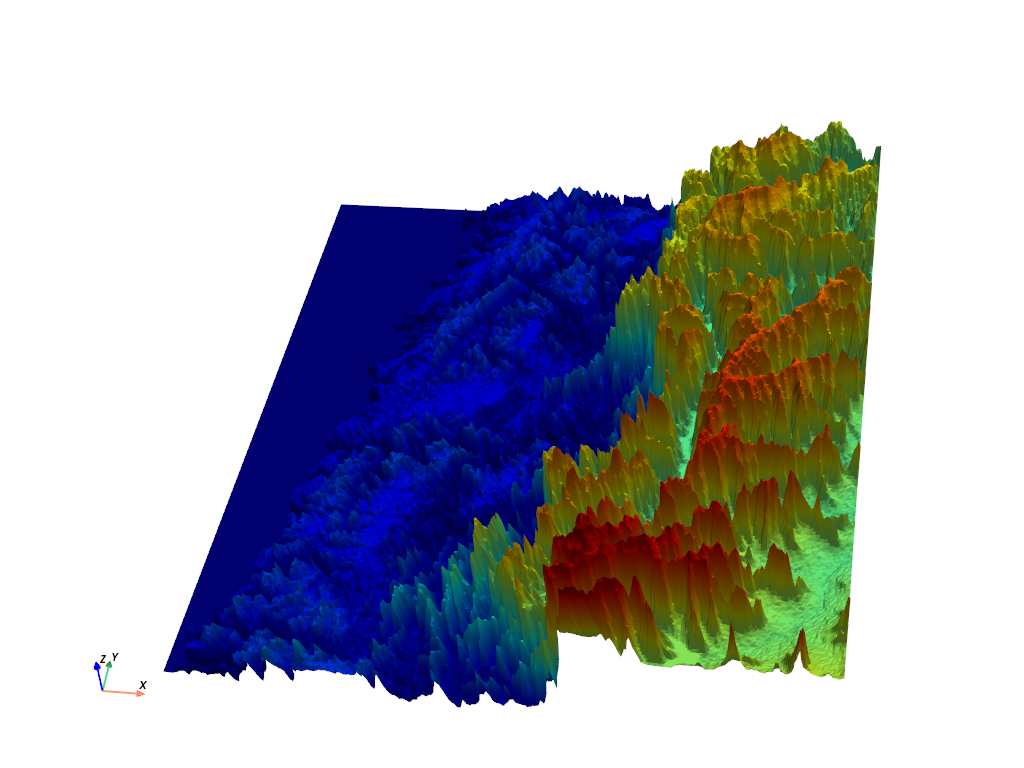}}
                                       &  SSGP++    &    8.40(0.45)e-02 &    -1.34(0.04)e+00 &    1.57(0.02)e+00\\
                                       && OVC       &    9.35(0.38)e-02 &    -1.13(0.01)e+00 &    1.18(0.01)e+00\\
                                       && OVC++     &    8.39(0.20)e-02 &    -1.30(0.01)e+00 & \B 8.53(0.12)e-01\\
                                       && POAM      & \B 7.73(0.28)e-02 & \B -1.46(0.04)e+00 & \B 8.43(0.06)e-01\\
                                       \midrule
        \multirow{4}{*}{\textbf{Env3}} &
        \multirow{4}{*}{\includegraphics[width=0.1\linewidth,trim={0 0 0 60},clip]{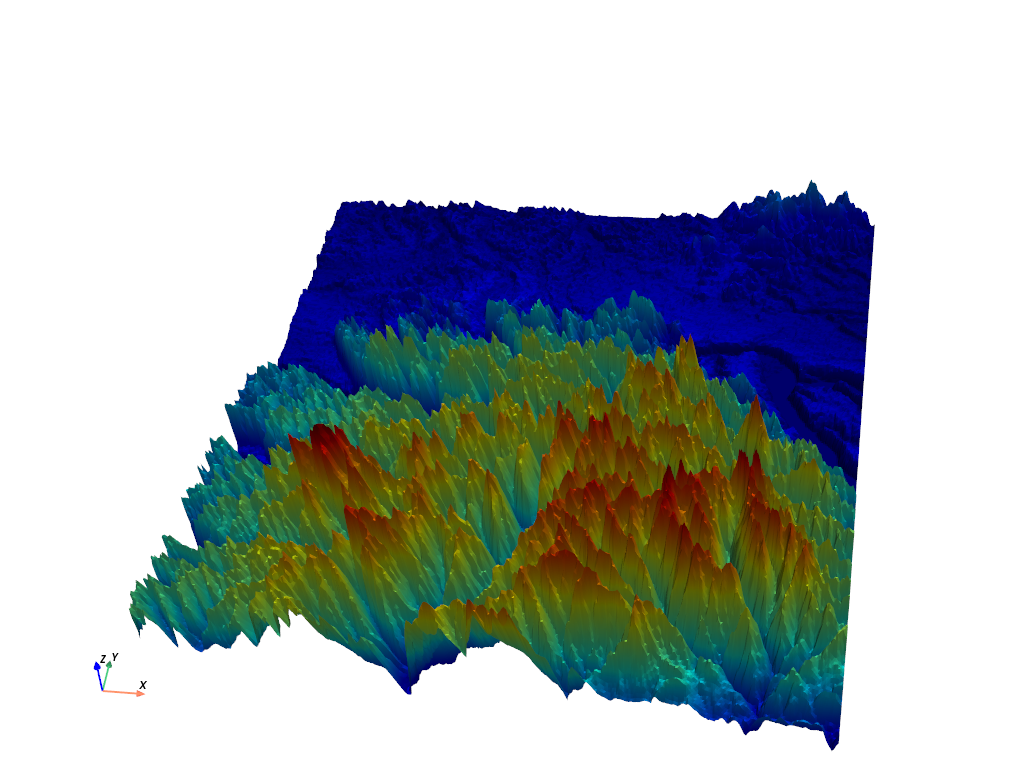}}
                                       &  SSGP++    &    1.68(0.11)e-01 &    -1.06(0.05)e+00 &    1.58(0.03)e+00\\
                                       && OVC       &    1.74(0.06)e-01 &    -8.82(0.13)e-01 &    1.18(0.01)e+00\\
                                       && OVC++     &    1.72(0.05)e-01 &    -1.03(0.04)e+00 & \B 8.50(0.12)e-01\\
                                       && POAM      & \B 1.62(0.09)e-01 & \B -1.14(0.02)e+00 & \B 8.46(0.11)e-01\\
                                       \midrule
        \multirow{4}{*}{\textbf{Env4}} &
        \multirow{4}{*}{\includegraphics[width=0.1\linewidth,trim={0 0 0 60},clip]{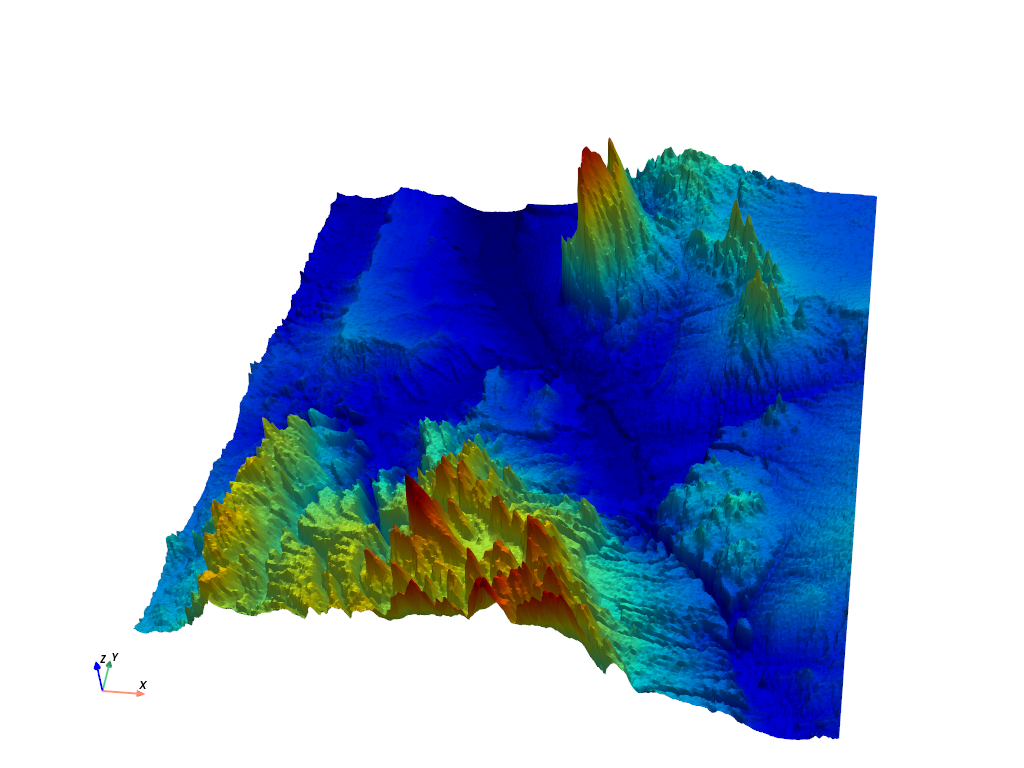}}
                                       &  SSGP++    &    9.79(2.34)e-02 &    -1.22(0.16)e+00 &    1.58(0.02)e+00\\
                                       && OVC       & \B 7.09(0.29)e-02 &    -1.28(0.02)e+00 &    1.18(0.01)e+00\\
                                       && OVC++     &    1.06(0.26)e-01 &    -1.15(0.14)e+00 & \B 8.59(0.10)e-01\\
                                       && POAM      &    8.30(1.16)e-02 & \B -1.42(0.05)e+00 & \B 8.54(0.07)e-01\\
                                       \bottomrule
    \end{tabular}%
\end{table*}

\begin{figure*}[tb]
  \centering
  \subfloat{{\includegraphics[width=0.5\linewidth]{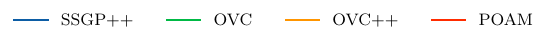}}}\vspace{-1em}\\
  \subfloat[SMSE in Env1]{\includegraphics[width=0.24\linewidth,height=0.13\linewidth,trim={0 4 0 1},clip]{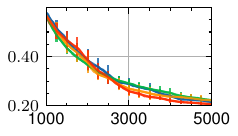}}%
  \subfloat[SMSE in Env2]{\includegraphics[width=0.24\linewidth,height=0.13\linewidth,trim={0 4 0 1},clip]{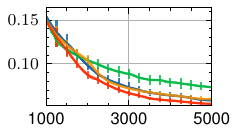}}%
  \subfloat[SMSE in Env3]{\includegraphics[width=0.24\linewidth,height=0.13\linewidth,trim={0 4 0 1},clip]{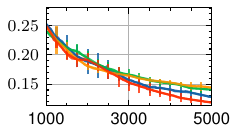}}%
  \subfloat[SMSE in Env4]{\includegraphics[width=0.24\linewidth,height=0.13\linewidth,trim={0 4 0 1},clip]{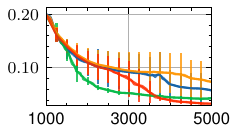}}\vspace{-1em}\\
  \subfloat[MSLL in Env1]{\includegraphics[width=0.24\linewidth,height=0.13\linewidth,trim={0 4 0 1},clip]{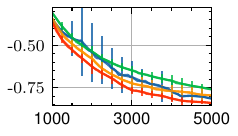}}%
  \subfloat[MSLL in Env2]{\includegraphics[width=0.24\linewidth,height=0.13\linewidth,trim={0 4 0 1},clip]{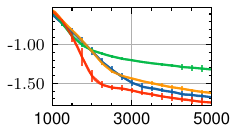}}%
  \subfloat[MSLL in Env3]{\includegraphics[width=0.24\linewidth,height=0.13\linewidth,trim={0 4 0 1},clip]{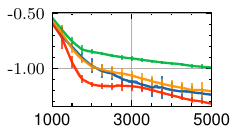}}%
  \subfloat[MSLL in Env4]{\includegraphics[width=0.24\linewidth,height=0.13\linewidth,trim={0 4 0 1},clip]{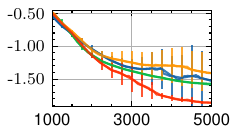}}\vspace{-1em}\\
  \subfloat[Time in Env1]{\includegraphics[width=0.24\linewidth,height=0.13\linewidth,trim={0 4 0 1},clip]{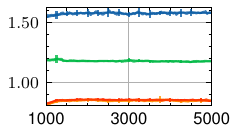}}%
  \subfloat[Time in Env2]{\includegraphics[width=0.24\linewidth,height=0.13\linewidth,trim={0 4 0 1},clip]{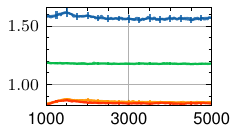}}%
  \subfloat[Time in Env3]{\includegraphics[width=0.24\linewidth,height=0.13\linewidth,trim={0 4 0 1},clip]{n44w111_train_time}}%
  \subfloat[Time in Env4]{\includegraphics[width=0.24\linewidth,height=0.13\linewidth,trim={0 4 0 1},clip]{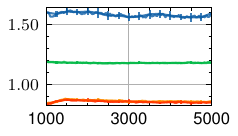}}\\
  \caption{\textbf{Benchmarking results of the evaluated methods across four environments and three metrics}. The proposed method (POAM) is compared against an online sparse GP baseline (\texttt{OVC}) and two improved baselines (\texttt{OVC++} and \texttt{SSGP++}) in terms of standardized mean squared error (SMSE), mean standardized log loss (MSLL), and training time.}\label{fig:quantitative}%
\end{figure*}

\begin{figure*}[tb]
    \centering
    \subfloat[SSGP++]{\includegraphics[width=0.23\linewidth,height=0.15\linewidth,trim={80 30 130 20},clip]{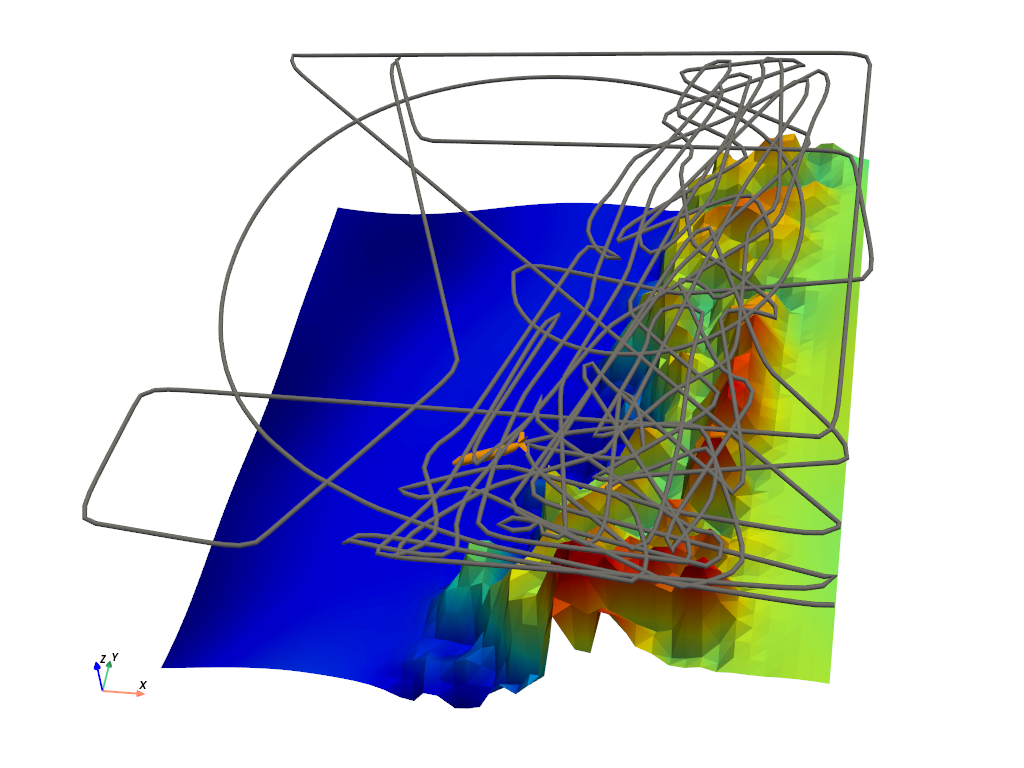}}%
    \subfloat[OVC]{\includegraphics[width=0.23\linewidth,height=0.15\linewidth,trim={80 30 130 20},clip]{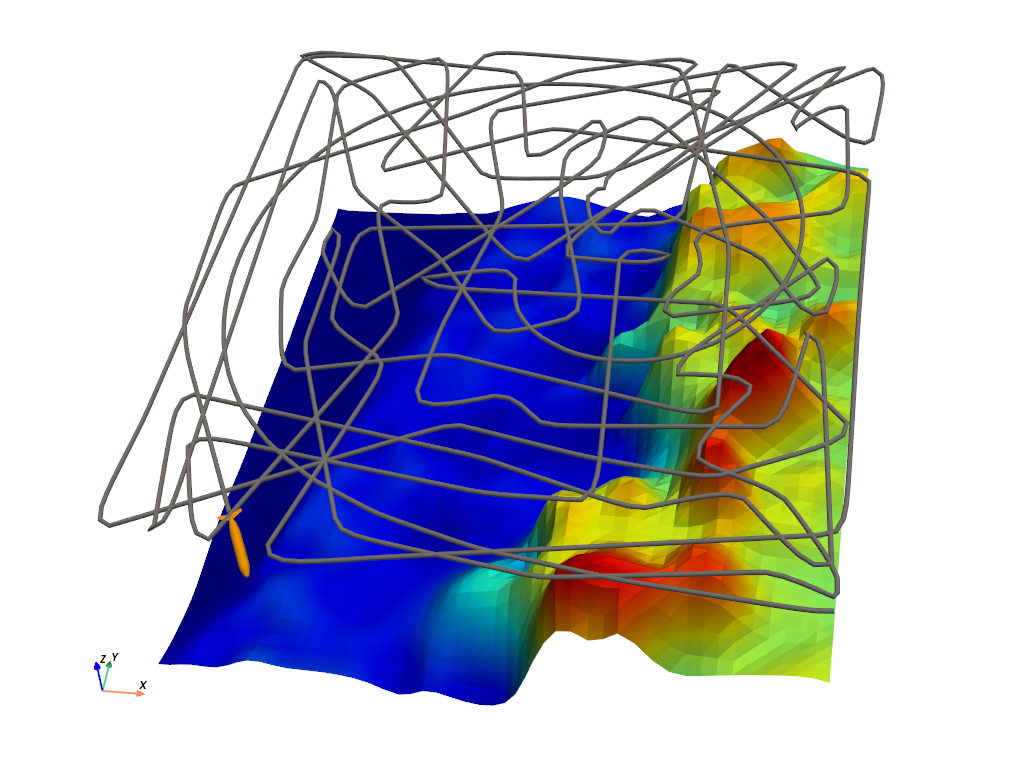}}%
    \subfloat[OVC++]{\includegraphics[width=0.23\linewidth,height=0.15\linewidth,trim={80 30 130 20},clip]{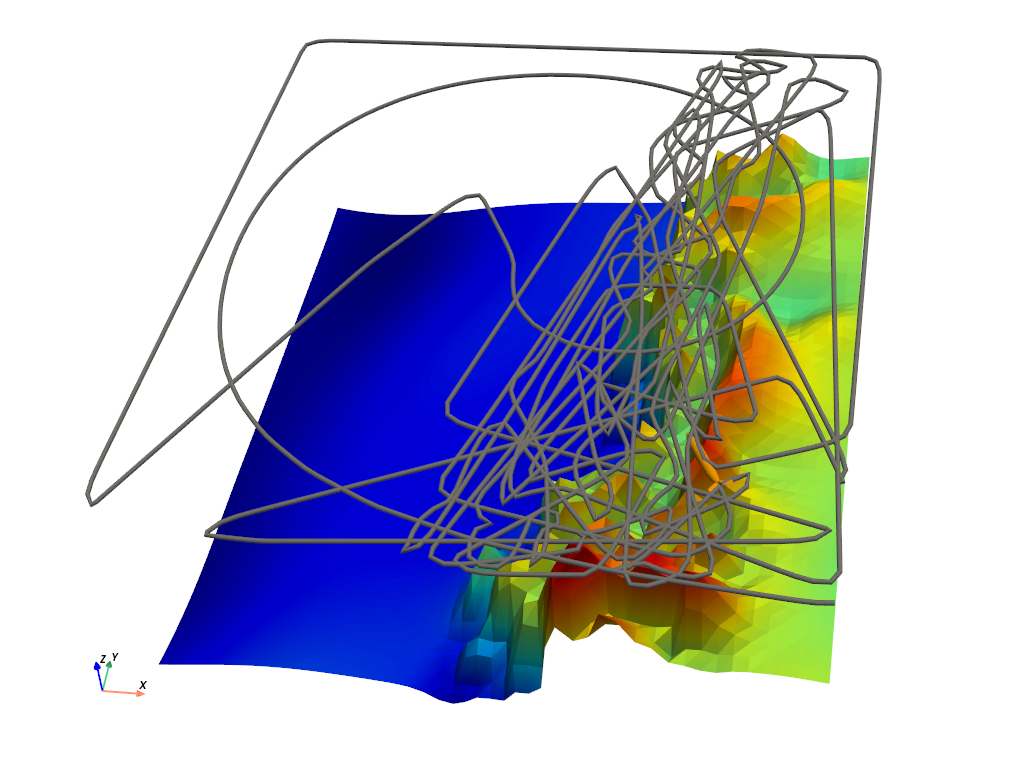}}%
    \subfloat[POAM]{\includegraphics[width=0.23\linewidth,height=0.15\linewidth,trim={80 30 130 20},clip]{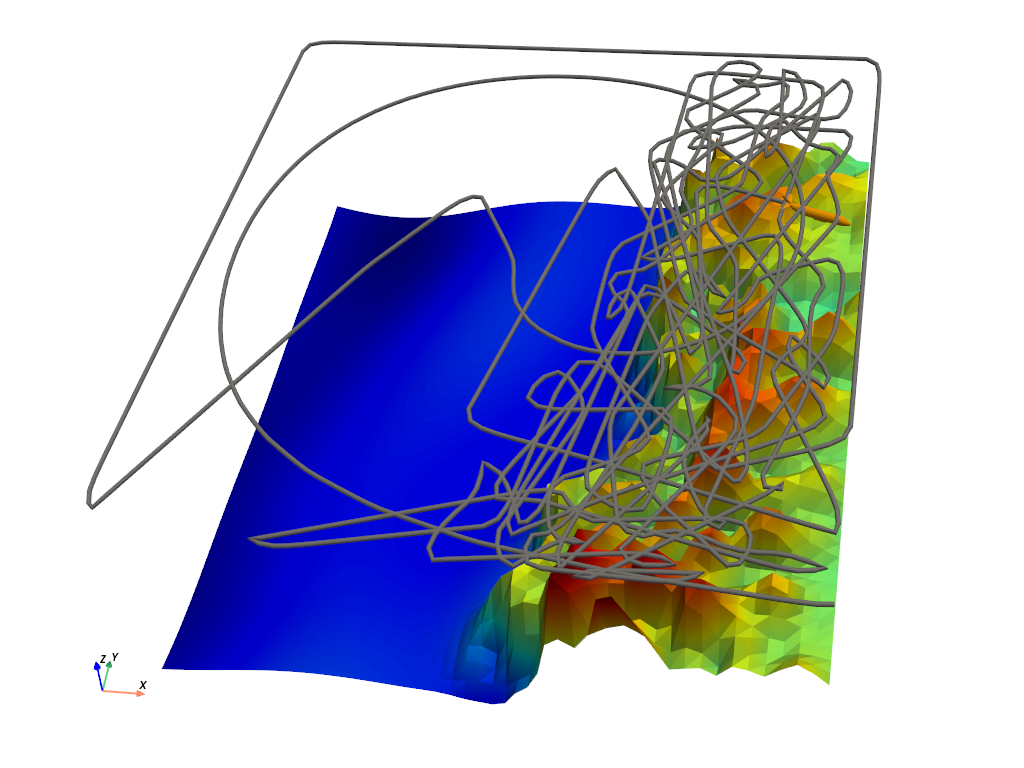}}    %
    \caption{\textbf{Prediction and sampling path of different methods in \texttt{Env2}}. The sampling paths of \texttt{SSGP++} and \texttt{OVC++} do not cover the right boundary extensively enough. \texttt{OVC} uniformly explores the environment. \texttt{POAM} effectively discovers and covers the complex region.}\label{fig:mean_path}
\end{figure*}

\begin{figure*}[tb]
    \centering
    \addtocounter{subfigure}{-15}%
    \subfloat{\raisebox{0.04\linewidth}{\rotatebox{90}{\textbf{SSGP++}}}}%
    \subfloat{\includegraphics[width=0.23\linewidth,height=0.15\linewidth,trim={80 25 80 80},clip]{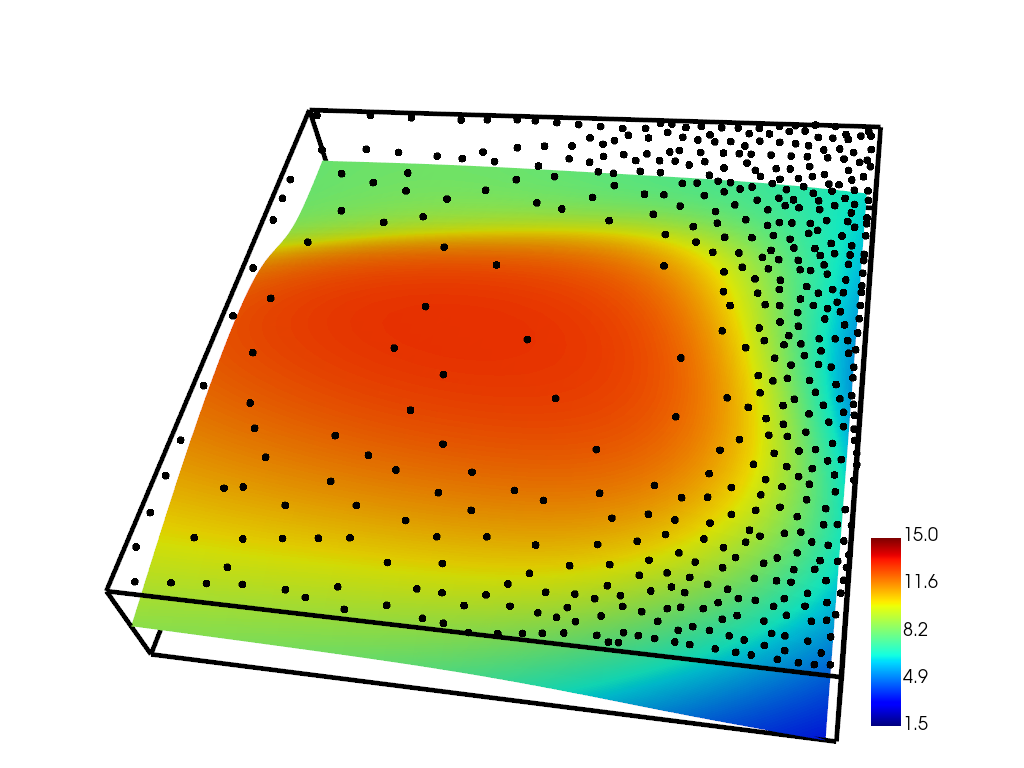}}%
    \subfloat{\includegraphics[width=0.23\linewidth,height=0.15\linewidth,trim={80 25 80 80},clip]{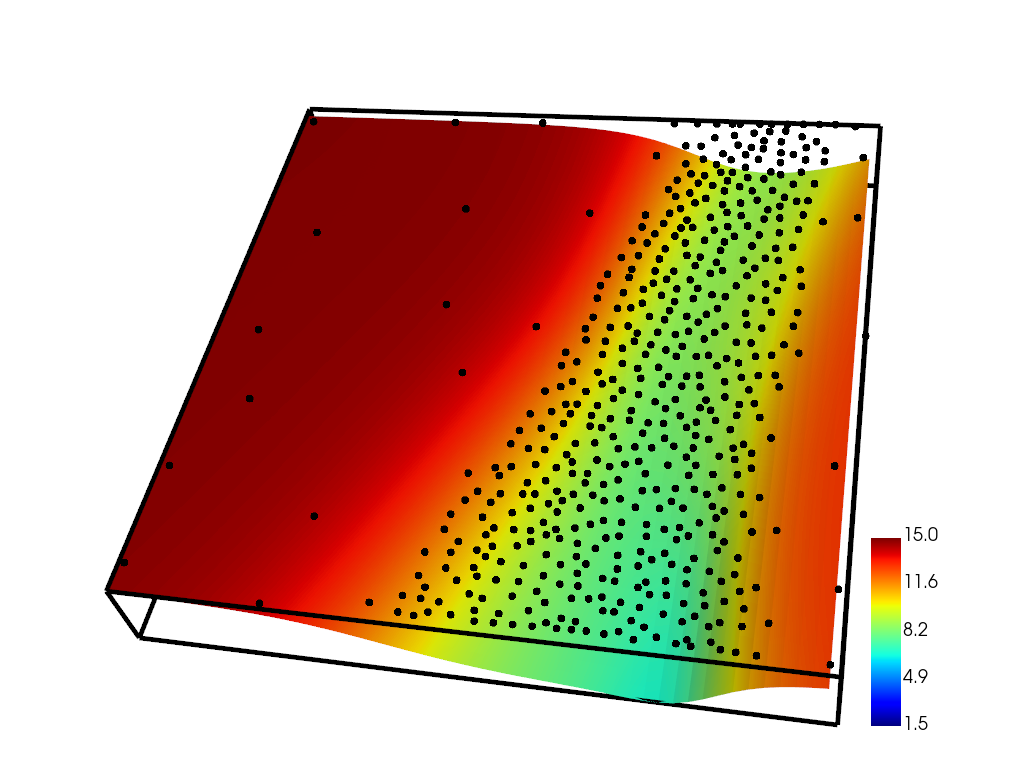}}%
    \subfloat{\includegraphics[width=0.23\linewidth,height=0.15\linewidth,trim={80 25 80 80},clip]{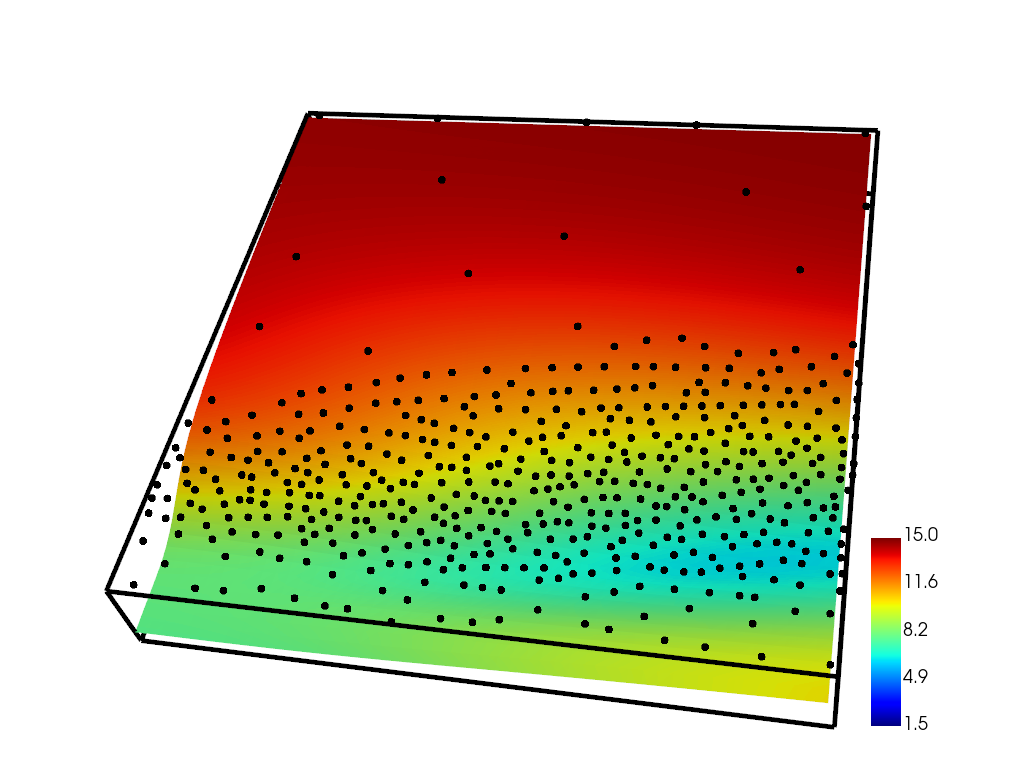}}%
    \subfloat{\includegraphics[width=0.23\linewidth,height=0.15\linewidth,trim={80 25 80 80},clip]{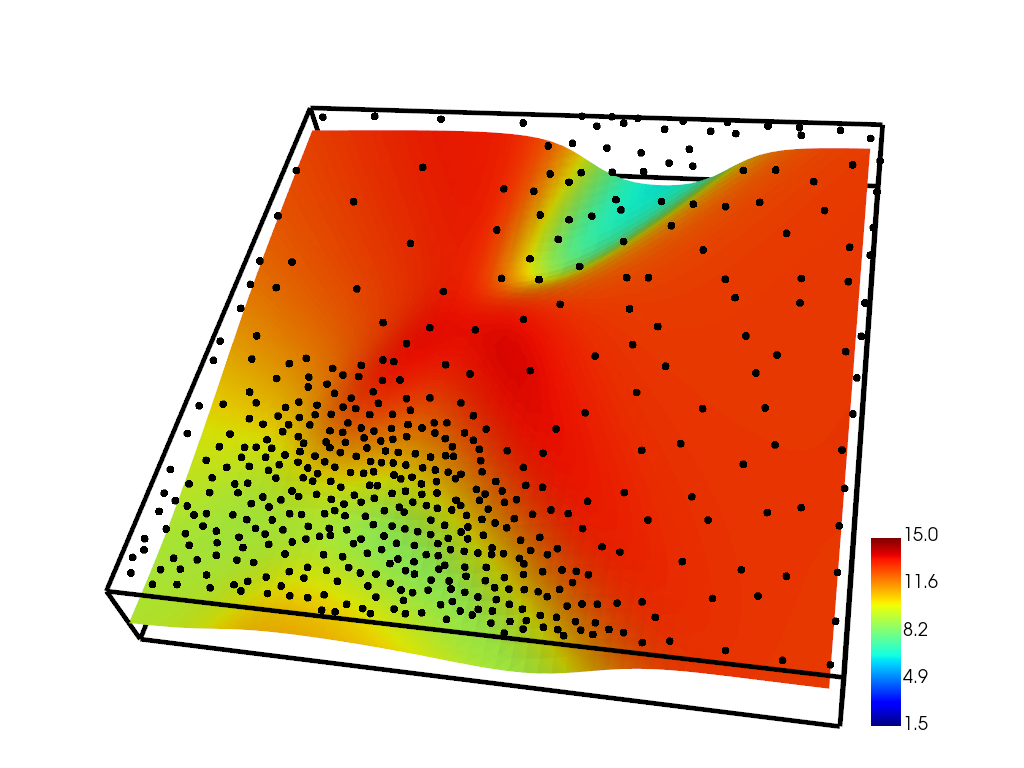}}\vspace{-1em}\\%
    \subfloat{\raisebox{0.04\linewidth}{\rotatebox{90}{\textbf{OVC++}}}}%
    \subfloat{\includegraphics[width=0.23\linewidth,height=0.15\linewidth,trim={80 25 80 80},clip]{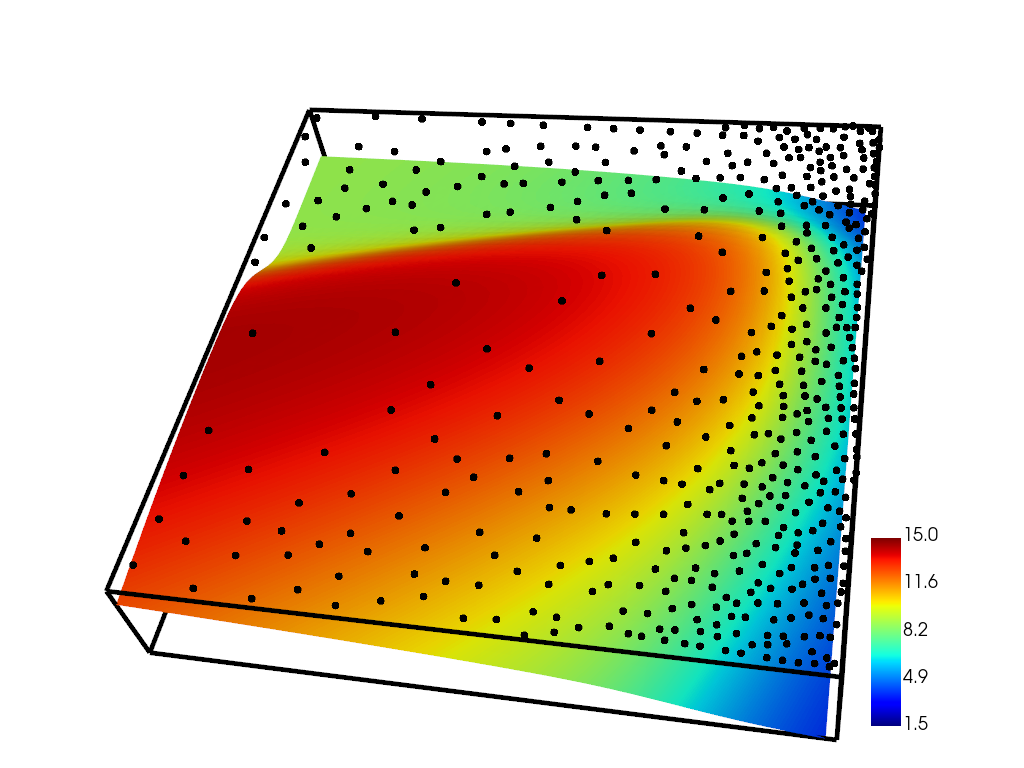}}%
    \subfloat{\includegraphics[width=0.23\linewidth,height=0.15\linewidth,trim={80 25 80 80},clip]{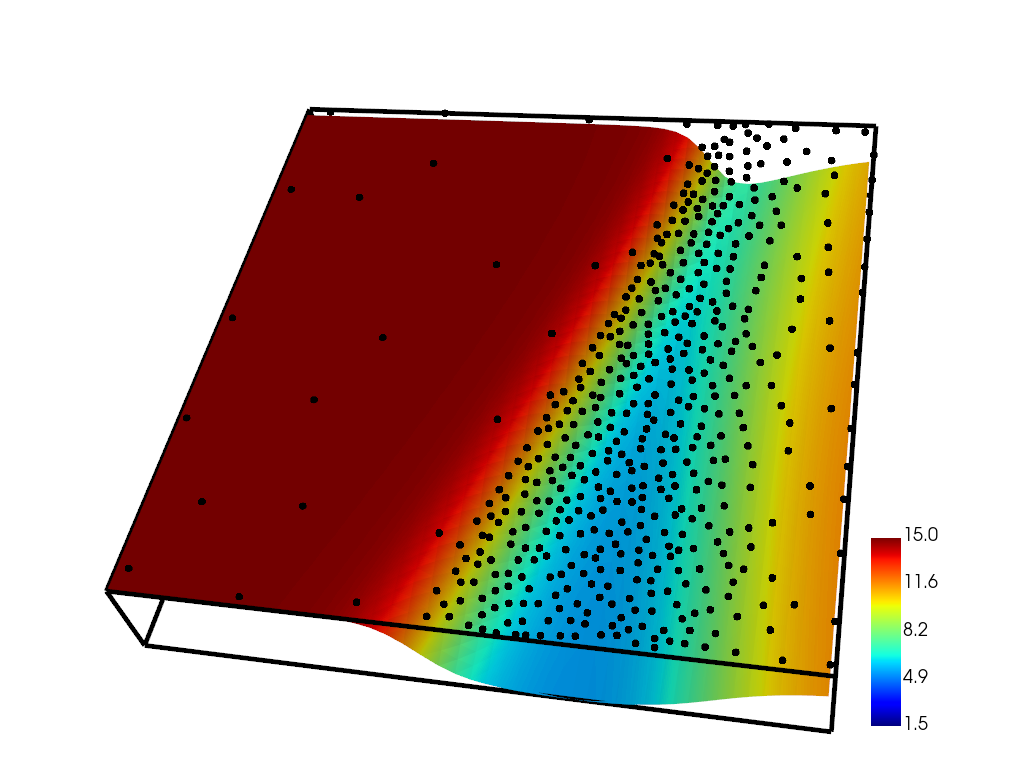}}%
    \subfloat{\includegraphics[width=0.23\linewidth,height=0.15\linewidth,trim={80 25 80 80},clip]{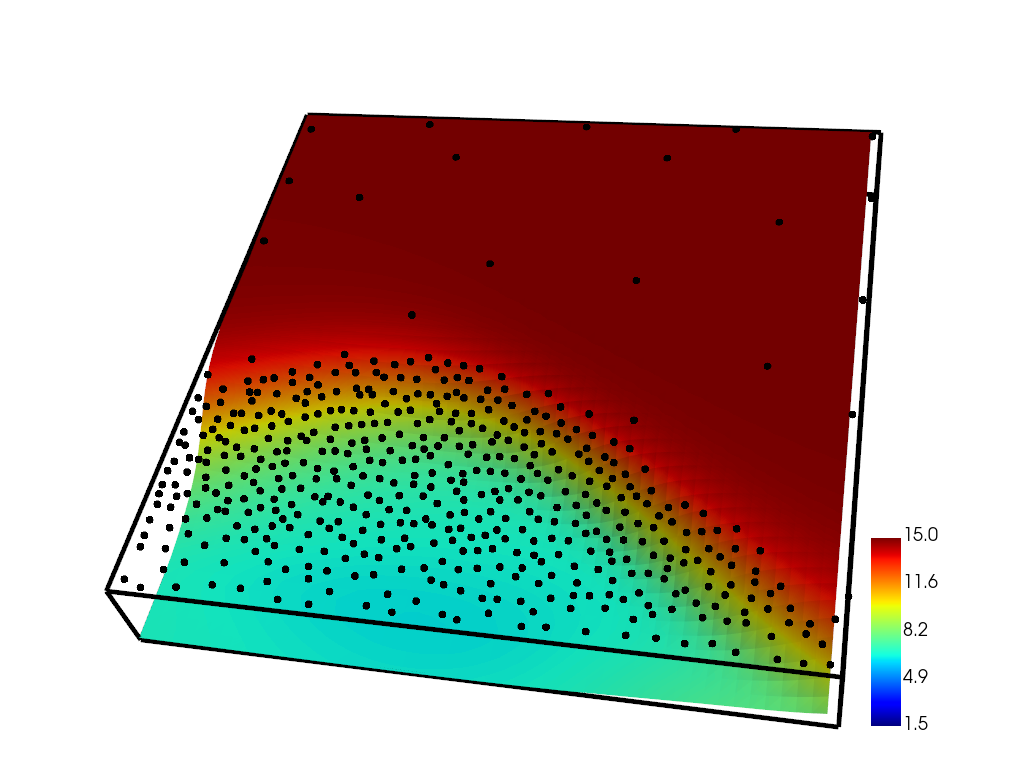}}%
    \subfloat{\includegraphics[width=0.23\linewidth,height=0.15\linewidth,trim={80 25 80 80},clip]{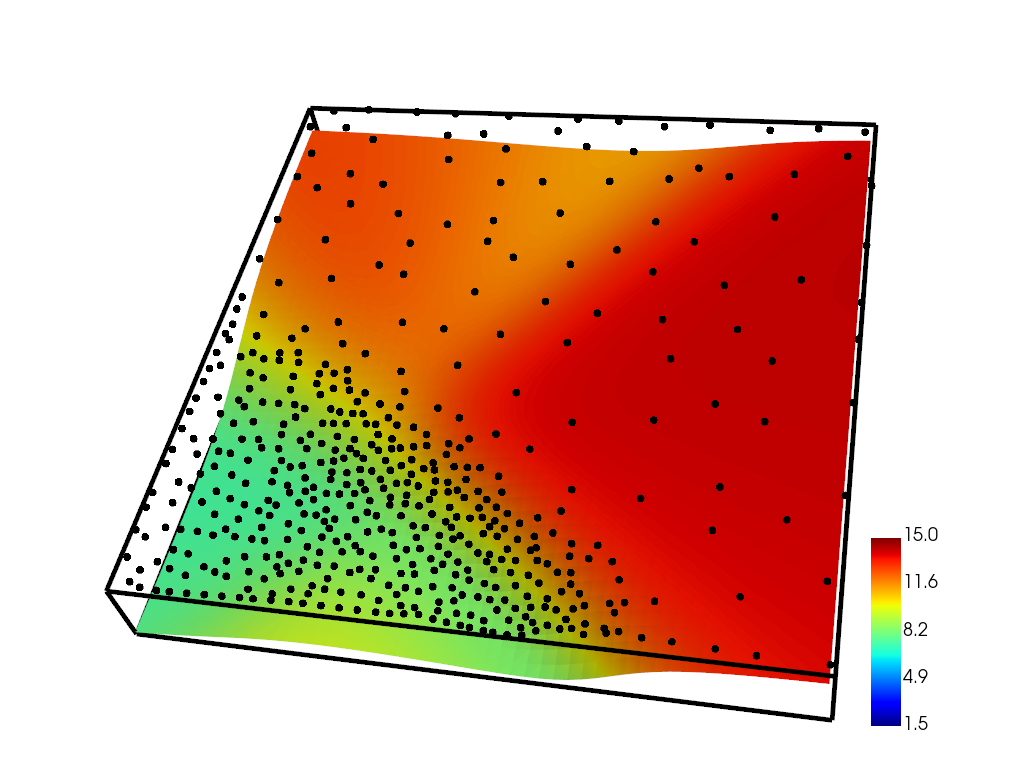}}\vspace{-1em}\\%
    \subfloat{\raisebox{0.04\linewidth}{\rotatebox{90}{\textbf{POAM}}}}%
    \subfloat{\includegraphics[width=0.23\linewidth,height=0.15\linewidth,trim={80 25 80 80},clip]{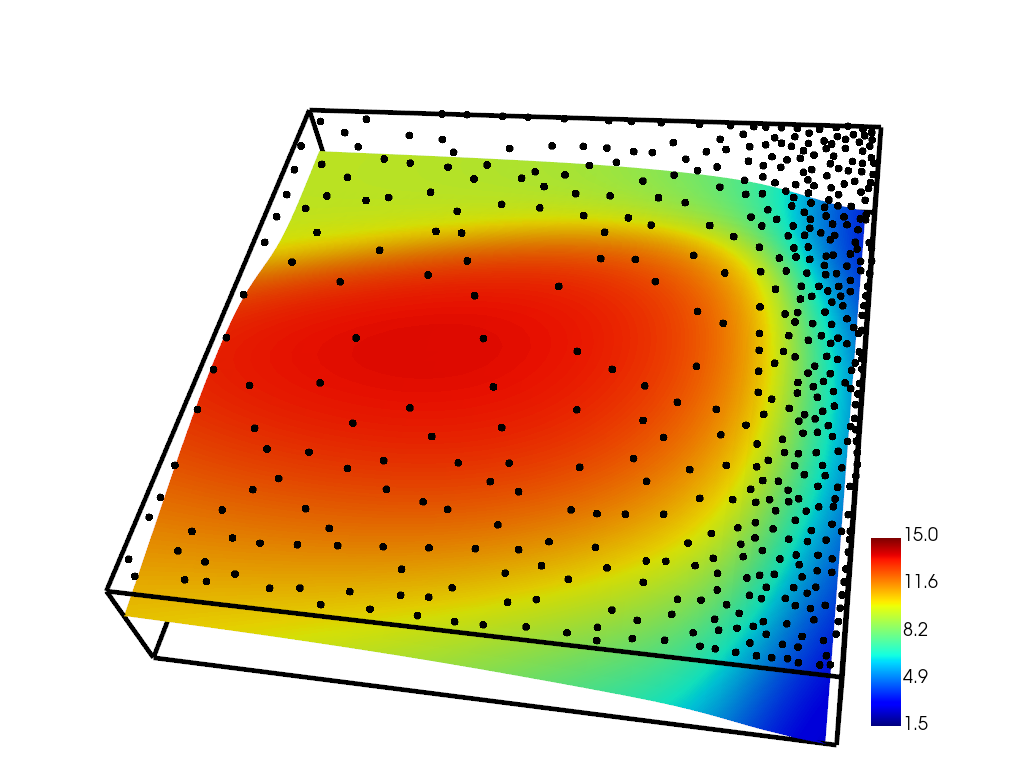}}%
    \subfloat{\includegraphics[width=0.23\linewidth,height=0.15\linewidth,trim={80 25 80 80},clip]{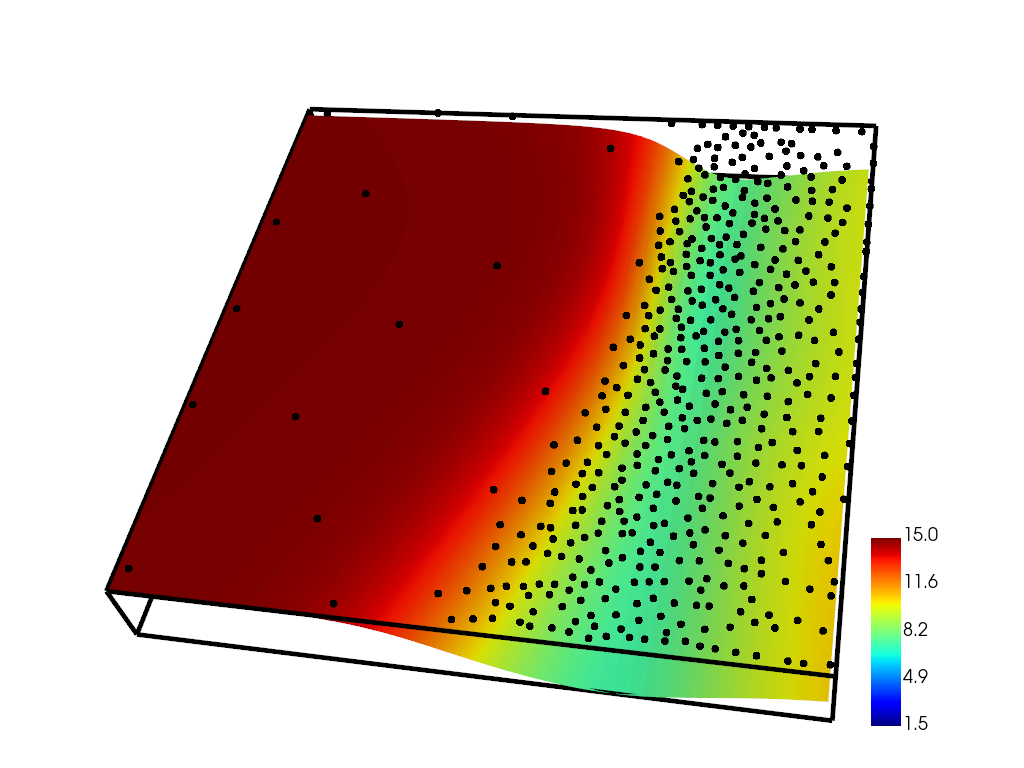}}%
    \subfloat{\includegraphics[width=0.23\linewidth,height=0.15\linewidth,trim={80 25 80 80},clip]{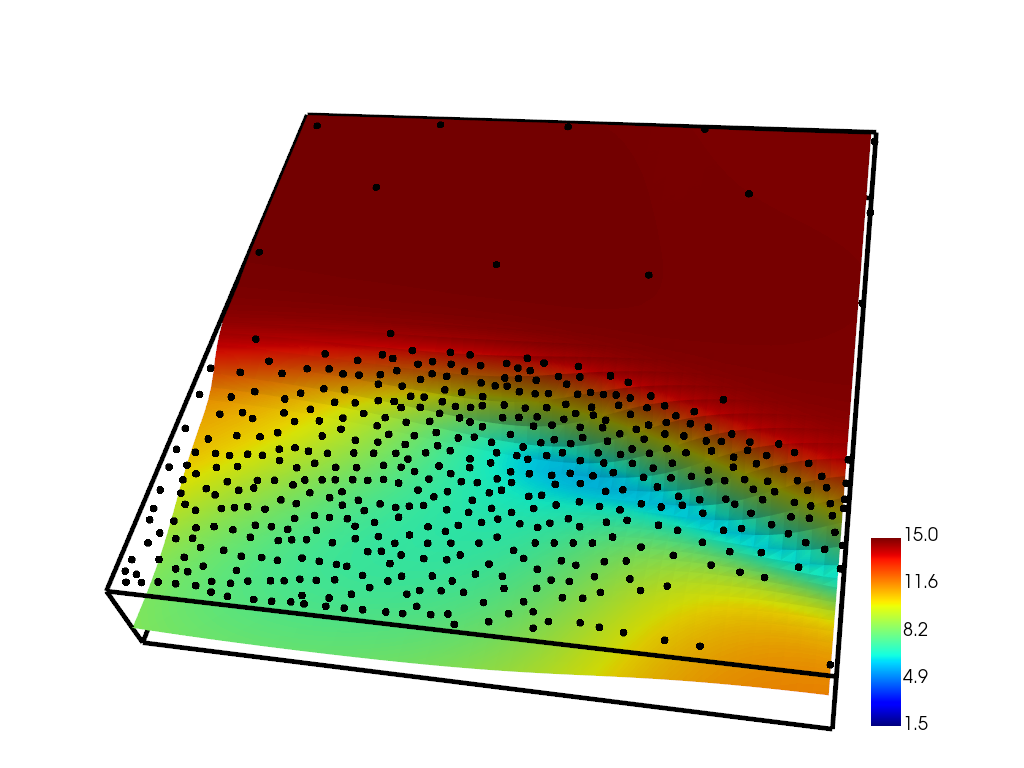}}%
    \subfloat{\includegraphics[width=0.23\linewidth,height=0.15\linewidth,trim={80 25 80 80},clip]{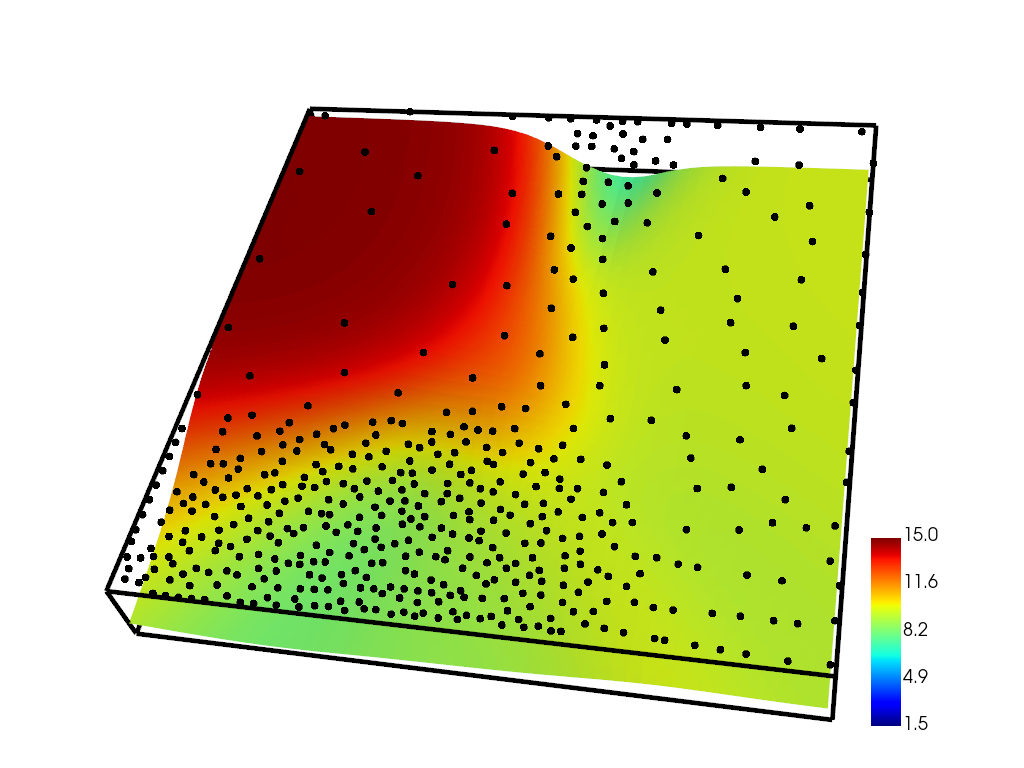}}\\%
    \subfloat[Env1\label{fig:env1}]{\includegraphics[width=0.23\linewidth,height=0.15\linewidth,trim={100 30 130 60},clip]{env_n44w111}}%
    \subfloat[Env2\label{fig:env2}]{\includegraphics[width=0.23\linewidth,height=0.15\linewidth,trim={100 30 130 120},clip]{env_n17e073}}%
    \subfloat[Env3\label{fig:env3}]{\includegraphics[width=0.23\linewidth,height=0.15\linewidth,trim={100 30 130 150},clip]{env_n47w124}}%
    \subfloat[Env4\label{fig:env4}]{\includegraphics[width=0.23\linewidth,height=0.15\linewidth,trim={100 30 130 120},clip]{env_n35w107}}
    \caption{\textbf{Visualization of the four environments and the learned lengthscale maps}. Red means high elevation while blue indicates low. The black dots represent the inducing points.}%
    \label{fig:lenscale_inducing}
\end{figure*}

\begin{figure*}[tb]
  \centering
  \subfloat{{\includegraphics[width=\linewidth]{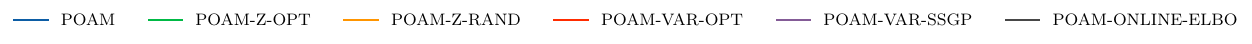}}}\vspace{-1em}\\
  \subfloat[SMSE]{\includegraphics[width=0.3\linewidth,height=0.15\linewidth,trim={0 4 0 1},clip]{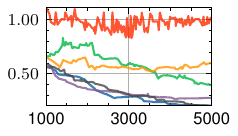}}%
  \subfloat[MSLL]{\includegraphics[width=0.3\linewidth,height=0.15\linewidth,trim={0 4 0 1},clip]{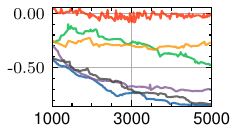}}%
  \subfloat[Time]{\includegraphics[width=0.3\linewidth,height=0.15\linewidth,trim={0 4 0 1},clip]{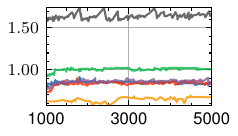}}%
  \caption{\textbf{Results of ablation study}. The experiment is conducted in Env1. Changing the inducing input update strategy to gradient-based optimization (\texttt{POAM-Z-OPT}) or random allocation (\texttt{POAM-Z-RAND}) deteriorates the performance. Updating the variational parameters using SSGP's strategy (\texttt{POAM-VAR-SSGP}) slightly reduces the performance. Training with the online evidence lower bound of SSGP (\texttt{POAM-ONLINE-ELBO}) does not change SMSE or MSLL but increases training time.}\label{fig:ablation}
\end{figure*}

\topic{Initial Sampling}
The robot initiates its trajectory by following a B\'ezier curve produced by 15 control points that adapt to the dimensions of the workspace, as illustrated in \Cref{fig:pilot_survey}. It is important to note that the pilot survey path is not limited to this B\'ezier curve, and the robot starting from the lower-right is an arbitrary choice; any initial path that helps the robot collect representative initial samples can be used.
Experiments showing robustness to different initialization strategies are discussed in Section~\ref{sub:robustness}.
These samples initialize the GP model and compute the statistics needed to normalize the input values to fall within the range of -1 to 1 and standardize the target values to have a mean of 0 and a standard deviation of 1. This preprocessing step enhances training efficacy and numerical stability, which is a common practice in GPR.

\topic{Environment Description}
We use digital elevation maps from the NASA Shuttle Radar Topography Mission (SRTM)~\cite{farr2007shuttle} to simulate the ground-truth environments for bathymetric mapping\footnote{The elevation map can be viewed and downloaded from the 30-Meter SRTM Tile Downloader \href{https://dwtkns.com/srtm30m/}{https://dwtkns.com/srtm30m/}.}.  The workspace dimensions are defined as $31\times{31}$ meters. For better visual comparison, the elevation maps of the four environments are shown later in \Cref{fig:lenscale_inducing}. Thumbnails of these environments are also provided in the second column of \Cref{tab:quantitative} for easier reference. These environments have distinct elevation patterns, which helps evaluate the performance of the compared methods across diverse scenarios, offering insights into their generalization capabilities.
\begin{itemize}
    \item \textbf{Env1}: The topography is relatively smooth on the left side and the center, but rocky in the other three directions.
    \item \textbf{Env2}: The terrain is relatively flat in the left two-thirds and mountainous in the remaining one-third.
    \item \textbf{Env3}: There is high variability in the lower part and small variation in the upper section.
    \item \textbf{Env4}: It has two rugged regions at the top and bottom, and a flat area in the upper-left corner.
\end{itemize}

\topic{Compared Methods}
As discussed in \Cref{sub:pam}, optimizing the locations of inducing points while learning an input-dependent lengthscale in the AK via gradient optimization is challenging. Consequently, the vanilla Streaming Sparse GP~(SSGP)~\cite{bui2017streaming} and the original Online Variational Conditioning~(OVC)~\cite{maddox2021conditioning} methods do not perform well. To make these baseline methods comparable, we apply the same PCD strategy for selecting the inducing inputs for SSGP and OVC, without further gradient optimization. These improved baselines are referred to as SSGP++ and OVC++. All methods utilize 500 inducing points and are optimized with 10 gradient steps per decision epoch. We keep all other settings unchanged, modifying only the update strategies for the inducing inputs, variational parameters, and hyperparameters.

\topic{Evaluation Metrics}
Following standard practice in the GP literature~\cite{rasmussen2005mit}, we use Standardized Mean Squared Error~(SMSE) and Mean Standardized Log Loss~(MSLL) to evaluate model accuracy and uncertainty quantification. SMSE is calculated as the mean squared error divided by the variance of the test targets. After standardization, a simple method that predicts using the mean of the training targets will have an SMSE of approximately 1. MSLL standardizes the log loss by subtracting the log loss of a naive model that predicts using a Gaussian distribution with the mean and variance of the training targets. An MSLL value close to zero indicates a naive method, while negative values indicate better performance. Additionally, we compare the training time to assess computational efficiency. Since all methods are variants of SVGP, prediction times are similar and thus not included in the comparison.

\subsection{Benchmarking Results}%
\label{sub:results}
\Cref{tab:quantitative} shows that POAM consistently outperforms the baselines in terms of averaged accuracy (SMSE), uncertainty quantification (MSLL), and computational efficiency (training time) in the first three environments. In the fourth environment, OVC achieves a better SMSE than POAM, while POAM still excels in MSLL and training time. The superior SMSE of OVC in this environment is due to its joint optimization of the inducing inputs and lengthscale via stochastic gradient, resulting in a constant lengthscale across the spatial domain. This causes GPR to assign high uncertainty to data-scarce areas, disregarding the complexity of the underlying function, and guiding the robot to explore the environment uniformly, thus increasing the likelihood of discovering the two isolated rugged regions. In contrast, the other three methods learn input-dependent lengthscales. After discovering one rugged region, these models assign high uncertainty to the complex region, causing the robot to stay and collect more samples until the uncertainty in that area is significantly reduced. This delayed discovery of the other rugged region results in a slower decrease in SMSE.

\Cref{fig:quantitative} shows the changes in SMSE, MSLL, and training time as the robot collects between 1000 and 5000 samples. POAM exhibits the most rapid decline in both SMSE and MSLL across all environments except the fourth. In Env4, the SMSE curve for POAM initially shows a higher SMSE than OVC and decreases more slowly. However, after collecting around 3000 samples, POAM quickly catches up and surpasses OVC when sampling in the second rugged region.
Thus while \Cref{tab:quantitative} shows that, when averaging over decision epochs, the SMSE of OVC in Env4 is lower, the final SMSE value of POAM is better.
SSGP++, OVC++, and POAM have higher standard deviations in SMSE and MSLL curves because their performance depends on when the robot encounters rugged regions. Notably, POAM's MSLL curve is significantly lower than those of other methods, indicating better uncertainty quantification. The myopic planner used in the experiments focuses only on a nearby waypoint with the highest entropy, making it challenging to guide the robot to other important regions after discovering the first rugged region unless the uncertainty in the first region is significantly reduced. An advanced informative planner could better utilize POAM's uncertainty quantification capability to improve performance.

The training time curves consistently demonstrate constant runtimes across all methods. POAM and OVC++ have the fastest runtimes, closely followed by OVC. SSGP++ has the slowest runtime due to the complex online ELBO computation. The vanilla SSGP can be more time-consuming because it optimizes all parameters until convergence, whereas SSGP++ only takes several gradient steps. OVC++ has a runtime nearly identical to POAM, as these methods are similar, differing only in what they cache. POAM saves variables related to the training data, while OVC++ saves \emph{noise-weighted} data-dependent variables. This subtle difference significantly enhances POAM's accuracy and uncertainty quantification capabilities.

The most substantial performance disparity among the four environments is evident in Env2, which is therefore used for a qualitative assessment. \Cref{fig:mean_path} shows the prediction and sampling trajectories of the four methods. Notably, OVC demonstrates uniform exploration throughout the environment because it faces challenges in learning an input-dependent length scale for AK. In contrast, the other three methods guide the robot to collect more samples in the mountainous region. Additionally, SSGP++ and OVC++ ignore the complex plateau at the right border and primarily gather samples along the steep slope. POAM captures the plateau comprehensively, collecting samples in a more balanced manner.

\Cref{fig:lenscale_inducing} presents the learned lengthscale and inducing inputs across all environments. OVC's lengthscale maps are excluded because they remain flat in all environments. Comparing the lengthscale maps in Env1 to those learned with dense full data in \Cref{fig:pam_lenscale}, we observe that all three methods appropriately learn the input-dependent lengthscale from sequentially received data batches.

While the lengthscale maps appear similar in Env1, the learned lengthscale maps differ in the other three environments. In Env2, SSGP++ and OVC++ show a large lengthscale in the far-right region, whereas POAM assigns a smaller lengthscale to the same area. This results in SSGP++ and OVC++ allocating low uncertainty and fewer inducing inputs in this region, leading to sparser sampling near the right border. In Env3, all three methods assign small length scales in the lower part where the terrain variation is evident. However, POAM delineates the boundary between the two regions more clearly. The most significant differences in lengthscale maps are observed in Env4. While all methods capture the rugged region at the bottom, OVC++ does not place a small lengthscale in the upper rugged region. Additionally, SSGP++ and OVC++ consider the right part of the environment to be smooth, while POAM considers the upper-left corner to be smooth.

\subsection{Ablation Study}%
\label{sub:ablation_study}
To evaluate the effectiveness of each component in our method, we conducted an ablation study by examining various configurations:
\begin{itemize}
  \item \texttt{POAM-Z-OPT}: Optimizes inducing inputs using gradients instead of PCD.
  \item \texttt{POAM-Z-RAND}: Selects inducing inputs randomly rather than using PCD.
  \item \texttt{POAM-VAR-OPT}: Optimizes variational parameters using gradients instead of computing them analytically.
  \item \texttt{POAM-VAR-SSGP}: Replaces the variational parameter update rules of POAM with those of SSGP.
  \item \texttt{POAM-ONLINE-ELBO}: Trains hyperparameters using the online ELBO of SSGP rather than the standard ELBO.
\end{itemize}

\Cref{fig:ablation} shows the results of the ablation study. Changing any part of POAM leads to a noticeable drop in performance, except for \texttt{POAM-ONLINE-ELBO}, which performs similarly to POAM but is more computationally expensive. The results for \texttt{POAM-Z-OPT} and \texttt{POAM-VAR-OPT} show that using analytical methods for inducing inputs and variational parameters is better than using gradient-based optimization. Compared to \texttt{POAM-Z-RAND}, we see that pivoted Cholesky decomposition is essential for POAM's performance. The result of \texttt{POAM-VAR-SSGP} indicates that POAM's method for updating variational parameters is more effective than the SSGP method.

\subsection{Robustness}%
\label{sub:robustness}

The appendix include additional experiments that explore the robustness of the algorithm and experimental setting. In particular, we repeat the experiments with a different initialization strategy that uses random points instead of the B\'ezier curve, and with an uninformed planner that chooses waypoints randomly. The experiments show that the conclusions given above are valid and hence robust to such modifications. In addition, we include a baseline that uses the full dataset for updates of variational parameters. The results show that the performance of POAM is very close to that of using the full dataset, indicating it provides an effective approximation method.

%
%
%
%
%

\section{Conclusion}%
\label{sec:conclusion}
In this paper, we developed the Probabilistic Online Attentive Mapping (POAM) framework to achieve computational efficiency and data efficiency simultaneously in robotic information gathering in non-stationary environments. 
This is achieved by constant-time model updates and variational Expectation-Maximization training of sparse Gaussian process regression with the Attentive Kernel.  Extensive experiments in active bathymetric mapping tasks show that POAM outperforms existing online sparse Gaussian Process models in terms of accuracy, uncertainty quantification, and efficiency. 

Despite the promising results, POAM has limitations. The recursive update of the inducing inputs may lead to suboptimal performance in streaming time-series data because some old inducing inputs will be removed in order to make room for new ones, which is a common issue in online sparse GPs. The online update expressions for the variational parameters are for regression tasks. Other tasks such as classification requires different update rules. The hyperparameters are updated slowly, which may not be optimal when prompt hyperparameter adaptation is needed. Addressing these limitations provides opportunities for future work. Additionally, POAM can be extended to other active information gathering tasks such as active implicit surface mapping~\cite{liu2021active}, online active dynamics learning~\cite{buisson2020actively}, and online active perception for locomotion~\cite{muenprasitivej2024bipedal}.

\section*{Acknowledgements}
We acknowledge the support of NSF with grant numbers 2006886, 2047169 and 2246261. We appreciate the constructive comments from the anonymous conference reviewers, which have significantly improved this paper. This research was supported in part by Lilly Endowment, Inc., through its support for the Indiana University Pervasive Technology Institute.

\bibliographystyle{plainnat}
\bibliography{references}
\appendix

The appendix provides a detailed derivation and explanation of prior work and how POAM fits in that context, as well as additional experimental results showing the robustness of POAM and the experimental setup used in the main paper.

\subsection{Detailed Discussion of SSGP and OVC}

We provide detailed explanations on the updates of variational parameters and hyperparameters in the related work. Additionally, we describe how we adapt existing methods to serve as robust baselines for our proposed approach. We analyze various methods from the perspective of updating data-dependent variables to highlight their similarities and differences, positioning our work within the existing methods for online sparse Gaussian process regression. Consistent mathematical notations, as summarized in \Cref{tab:notations}, are used throughout the paper.

\begin{table}[tbp]
    \caption{Mathematical Notations.}\label{tab:notations}
    \centering\scriptsize
    \begin{tabular}{lll}
        \toprule
        {Meaning}           & {Example}           & {Remark}\\
        \midrule
        variable            & $n$                 & lower-case\\
        constant            & $N$                 & upper-case\\
        vector              & $\mathbf{x}$        & bold, lower-case\\
        matrix              & $\mathbf{X}$        & bold, upper-case\\
        set/space           & $\mathbb{R}$        & blackboard\\
        function            & $f(\cdot)$ & \\
        functional          & $\mathtt{KL}[\cdot]$     & typewriter with square brackets\\
        special density     & $\mathcal{N}$       & calligraphy capital\\
        definition          & $\triangleq$        & normal\\
        transpose           & $\mathbf{m}^{\T}$   & customized command\\
        Euclidean norm      & $\norm{\cdot}$      & customized command\\
        \bottomrule
    \end{tabular}%
\end{table}

\subsubsection{Sparse Gaussian Process Regression~(SGPR)}%
We reintroduce the predictive distribution of SGPR for easier reference:
\begin{align}
  q(f_{\star}) =& \mathcal{N}\left(f_{\star} \mid \tilde{\mu}, \tilde{\nu}\right)\notag\\
  \tilde{\mu} =& \mathbf{k}_{u \star}^{\T} \mathbf{K}_{u u}^{-1} \mathbf{m},\notag\\
  \tilde{\nu} =& k_{\star \star}-\mathbf{k}_{u \star}^{\T} \mathbf{K}_{u u}^{-1} \mathbf{k}_{u \star}+\mathbf{k}_{u \star}^{\T} \mathbf{K}_{u u}^{-1} \mathbf{S} \mathbf{K}_{u u}^{-1} \mathbf{k}_{u \star},\notag\\
  \mathbf{S} =& \mathbf{K}_{u u} \mathbf{A}^{-1} \mathbf{K}_{u u},\label{ap:sgpr_S}\\
  \mathbf{m} =& \mathbf{K}_{u u} \mathbf{A}^{-1} {\color{blue} \mathbf{K}_{u f} \mathbf{\Sigma}^{-1} \mathbf{y}},\label{ap:sgpr_m}\\
  \mathbf{A} =& \mathbf{K}_{u u} + {\color{red} \mathbf{K}_{u f} \mathbf{\Sigma}^{-1} \mathbf{K}_{u f}^{\T}}.\notag
\end{align}
Here, $\mathbf{k}_{u \star}$ is the kernel vector between the inducing inputs and the test inputs, $\mathbf{K}_{u u}$ is the kernel matrix of the inducing inputs, $\mathbf{K}_{u f}$ is the kernel matrix between the inducing inputs and the training inputs, and $\mathbf{\Sigma}$ is a diagonal matrix with the noise variance $\sigma^{2}$ on the diagonal.
We discuss different methods from the perspective of how they update the variational parameters $\mathbf{m}$ and $\mathbf{S}$ via the data-dependent terms, which are highlighted in {\color{blue} blue} (for the $\mathbf{y}$-related term) and {\color{red} red}.

\subsubsection{Streaming Sparse Gaussian Processes~(SSGP)}%
Streaming sparse GPs~(SSGP) extends SGPR to the streaming setting where data arrives sequentially and the model can only process each batch of data once before discarding it~\cite{bui2017streaming}.
To derive an online update that facilitates hyperparameter optimization,
SSGP approximates the likelihood of the old data with the old posterior $q'(\mathbf{u'})$ and prior $p'(\mathbf{u'})$ at the old inducing inputs $\mathbf{u'}$.
This results in two additional KL regularization terms in the ELBO:
\begin{align}
  \mathtt{ELBO} =& \sum_{n=1}^{N_{\text{new}}}\mathtt{E}_{q(f_{n})}\left[\log p(y_{n} \mid f_{n})\right] - \mathtt{KL}[q(\mathbf{u})\ \|\ p(\mathbf{u})]\notag\\
  +& \mathtt{KL}[q(\mathbf{u'})\ \|\ p'(\mathbf{u'})] - \mathtt{KL}[q(\mathbf{u'})\ \|\ q'(\mathbf{u'})].\label{eq:ssgp_uncollapsed_elbo}
\end{align}
The first line of \Cref{eq:ssgp_uncollapsed_elbo} is the uncollapsed ELBO of SVGP in \Cref{eq:svgp_elbo} applied to the new data, and the second line involves two KL regularization terms for online update.

As in SGPR, this generic ELBO can be further simplified in the regression case to a collapsed form:
\begin{align}
  \label{eq:ssgp_elbo}
  \mathtt{ELBO} =& \log\mathcal{N}(\mathbf{\tilde{y}} \mid \mathbf{0}, \mathbf{Q}_{\tilde{y}\tilde{y}}) - \frac{1}{2\sigma^{2}}\mathrm{tr}(\mathbf{K}_{ff}-\mathbf{Q}_{ff}) \notag\\
                 &- \frac{1}{2}\mathrm{tr}(\mathbf{\hat{\Sigma}}^{-1}(\mathbf{K}_{u' u'} - \mathbf{Q}_{u' u'})) + \text{constants}.
\end{align}
We follow appendix C.4 of \citet{maddox2021conditioning} and collect variables that are not related to the trainable parameters to the \texttt{constants} term.
Here, some notations are introduced to reveal the similarity and difference between the collapsed ELBO of SSGP and that of SGPR:
\begin{align}
  \mathbf{\hat{\Sigma}}=(\mathbf{S'}^{-1}-\mathbf{K'}_{u' u'}^{-1})^{-1},&\qquad
  \mathbf{Q}_{u' u'}=\mathbf{K}_{u u'}^{\T}\mathbf{K}_{u u}^{-1}\mathbf{K}_{u u'},\notag\\
  \mathbf{\tilde{y}}=[\mathbf{\hat{y}}^{\T}, \mathbf{y}^{\T}]^{\T},&\qquad
  \mathbf{\hat{y}} = \mathbf{\hat{\Sigma}}\mathbf{S'}^{-1}\mathbf{m'},\notag\\
  \mathbf{Q}_{\tilde{y}\tilde{y}}=\mathbf{Q}_{\tilde{f} \tilde{f}} + \bm{\tilde{\Sigma}},&\qquad
  \mathbf{Q}_{\tilde{f} \tilde{f}} = \mathbf{K}_{u \tilde{f}}^{\T}\mathbf{K}_{u u}^{-1}\mathbf{K}_{u \tilde{f}},\notag\\
  \bm{\tilde{\Sigma}}=\begin{bmatrix}\mathbf{\hat{\Sigma}} & \mathbf{0} \\ \mathbf{0} & \bm{\Sigma}\end{bmatrix},&\qquad
  \mathbf{K}_{u \tilde{f}}=[\mathbf{K}_{u u'}, \mathbf{K}_{u f}].\label{eq:ssgp_notations}
\end{align}
The first line of \Cref{eq:ssgp_elbo} has a similar form as the collapsed ELBO of SGPR in \Cref{eq:collapsed_elbo} and the second trace term regularizes the new inducing inputs to be close to the old ones.
Other terms that are not related to the trainable parameters are represented as ``constants" in the equation above.
With this online ELBO, the old data is not needed and
hyperparameters and inducing inputs can be directly optimized using only new data.
To aid the optimization of the inducing inputs, \citet{bui2017streaming} also propose a resampling heuristic for initialization of inducing inputs.
Specifically, some randomly selected old inducing inputs are moved to randomly selected inputs in the new batch.

From Eq.\,(42) and Eq.\,(45) in \citet{bui2017streaming}'s appendix\footnote{We refer to equations in their revised arXiv paper available at \url{https://arxiv.org/abs/1705.07131}, noting that the equation numbering differs from the proceeding version.}, the predictive distribution is given by:
\begin{align*}
  q(f_{\star}) =& \mathcal{N}\left(f_{\star} \mid \tilde{\mu}, \tilde{\nu}\right),\\
  \tilde{\mu} =& \mathbf{k}_{u \star}^{\T} \mathbf{K}_{u u}^{-1} \mathbf{\tilde{m}},\\
  \tilde{\nu} =& k_{\star \star}-\mathbf{k}_{u \star}^{\T} \mathbf{K}_{u u}^{-1} \mathbf{k}_{u \star}+\mathbf{k}_{u \star}^{\T} \mathbf{K}_{u u}^{-1} \mathbf{\tilde{S}} \mathbf{K}_{u u}^{-1} \mathbf{k}_{u \star},\\
  \text{where }\mathbf{\tilde{S}} =& (\mathbf{K}_{u u}^{-1} + \mathbf{K}_{u u}^{-1} \mathbf{K}_{u \tilde{f}} \mathbf{\tilde{\Sigma}}^{-1} \mathbf{K}_{\tilde{f} u} \mathbf{K}_{u u}^{-1})^{-1},\\
\mathbf{\tilde{m}} =& \mathbf{\tilde{S}} \mathbf{K}_{u u}^{-1} \mathbf{K}_{u \tilde{f}} \mathbf{\tilde{\Sigma}}^{-1} \mathbf{\tilde{y}}.
\end{align*}
To see how the data-dependent terms are updated, we simplify the expressions of $\mathbf{\tilde{S}}$ to make it similar to SGPR's expression in \Cref{ap:sgpr_S}:
\begin{align}
  \mathbf{\tilde{S}} =& (\mathbf{K}_{u u}^{-1} + \mathbf{K}_{u u}^{-1} \mathbf{K}_{u \tilde{f}} \mathbf{\tilde{\Sigma}}^{-1} \mathbf{K}_{\tilde{f} u} \mathbf{K}_{u u}^{-1})^{-1}\notag\\
  =& {\color{gray} \mathbf{K}_{u u} \mathbf{K}_{u u}^{-1}} (\mathbf{K}_{u u}^{-1} + \mathbf{K}_{u u}^{-1} \mathbf{K}_{u \tilde{f}} \mathbf{\tilde{\Sigma}}^{-1} \mathbf{K}_{\tilde{f} u} \mathbf{K}_{u u}^{-1})^{-1} {\color{gray} \mathbf{K}_{u u}^{-1} \mathbf{K}_{u u}}\notag\\
  =& \mathbf{K}_{u u} {\underbracket{(\mathbf{K}_{u u} + {\color{red} \mathbf{K}_{u \tilde{f}} \mathbf{\tilde{\Sigma}}^{-1} \mathbf{K}_{\tilde{f} u}})}_{\mathbf{\tilde{A}}}}^{-1} \mathbf{K}_{u u}.\label{ap:ssgp_S}
\end{align}
Plugging the simplified $\mathbf{\tilde{S}}$ into $\mathbf{\tilde{m}}$, we have an $\mathbf{\tilde{m}}$ expression that resembles SGPR's $\mathbf{m}$ in \Cref{ap:sgpr_m}:
\begin{align}
  \mathbf{\tilde{m}}=\mathbf{K}_{u u} \mathbf{\tilde{A}}^{-1} {\color{blue}\mathbf{K}_{u \tilde{f}} \mathbf{\tilde{\Sigma}}^{-1} \mathbf{\tilde{y}}}.\label{ap:ssgp_m}
\end{align}
According to Eq.\,(33) and Eq.\,(38) in \citet{bui2017streaming}'s appendix, these data-dependent terms are expanded as
\begin{align}
  &{\color{blue}\mathbf{K}_{u \tilde{f}} \mathbf{\tilde{\Sigma}}^{-1} \mathbf{\tilde{y}}}\notag\\
  =& {\color{blue} \mathbf{K}_{u f} \mathbf{\Sigma}^{-1} \mathbf{y}} + \mathbf{K}_{u' u}^{\T} {\color{green} \mathbf{S'}^{-1} \mathbf{m'}},\label{ap:ssgp_c}\\
  &{\color{red}\mathbf{K}_{u \tilde{f}} \mathbf{\tilde{\Sigma}}^{-1} \mathbf{K}_{\tilde{f} u}}\notag\\
  =& {\color{red} \mathbf{K}_{u f} \mathbf{\Sigma}^{-1} \mathbf{K}_{f u}} + \mathbf{K}_{u' u}^{\T} ({\color{green} \mathbf{S'}^{-1} - \mathbf{K'}_{u' u'}^{-1}}) \mathbf{K}_{u' u}.\label{ap:ssgp_C}
\end{align}
Now we can see that SSGP updates the variational parameters $\mathbf{m}$ and $\mathbf{S}$ by saving the old variational parameters and kernel matrix (highlighted in {\color{green} green}), correcting/projecting them via the cross-covariance matrix $\mathbf{K}_{u' u}$, and adding the new data-dependent terms $\mathbf{K}_{u f} \mathbf{\Sigma}^{-1} \mathbf{y}$ and $\mathbf{K}_{u f} \mathbf{\Sigma}^{-1} \mathbf{K}_{f u}$.

\subsubsection{Online Variational Conditioning~(OVC)}%
Observing the similarity between \Cref{ap:sgpr_S,ap:ssgp_S} as well as \Cref{ap:sgpr_m,ap:ssgp_m},
\citet{maddox2021conditioning} view SSGP's variational update as a sequence of SGPR updates on an \emph{augmented} dataset $\{\mathbf{\tilde{X}}, \mathbf{\tilde{y}}\}$ that consists of pseudo data $\{\mathbf{Z'},\mathbf{\hat{y}}\}$ and the current data $\{\mathbf{X},\mathbf{y}\}$  through a special likelihood function $p(\mathbf{\tilde{y}}\,|\,\mathbf{\tilde{f}})=\mathcal{N}(\mathbf{\tilde{y}} \mid \mathbf{0}, \mathbf{\tilde{\Sigma}})$.
Specifically, the augmented covariance matrix $\bm{\tilde{\Sigma}}$ is a block-diagonal matrix with the pseudo covariance $\mathbf{\hat{\Sigma}}$ and the new covariance $\bm{\Sigma}$ on the diagonal, and the augmented dataset is defined as $\mathbf{\tilde{X}}=[\mathbf{Z'}^{\T},\mathbf{X}^{\T}]^{\T}, \mathbf{\tilde{y}}=[\mathbf{\hat{y}}^{\T},\mathbf{y}^{\T}]^{\T}$.
Intuitively, the information of the old data is preserved in the pseudo data $\{\mathbf{Z'},\mathbf{\hat{y}}\}$ and the pseudo covariance $\bm{\hat{\Sigma}}$.
In the following, variables related to the augmented dataset are denoted with a tilde symbol $\tilde{\cdot}$ while those related to the pseudo dataset are denoted with a hat $\hat{\cdot}$.

Computing the pseudo targets $\mathbf{\hat{y}}$ and the pseudo covariance matrix $\bm{\hat{\Sigma}}$ is the key step in constructing the augmented dataset.
\citet{maddox2021conditioning} show that  (\textit{cf.} Eq.\,(A.6) and Eq.\,(A.7)) they can be derived by reversing the equations of the old variational parameters $\mathbf{m'}$ and $\mathbf{S'}$ in \Cref{ap:sgpr_m,ap:sgpr_S} when assuming that the old data are observed only at the old inducing inputs $\mathbf{X'}=\mathbf{Z'}$:
\begin{align}
  &\mathbf{S'} = \mathbf{K'}_{u' u'} (\mathbf{K'}_{u' u'} + \mathbf{K'}_{u' u'} \mathbf{\hat{\Sigma}}^{-1} \mathbf{K'}_{u' u'})^{-1} \mathbf{K'}_{u' u'}\notag\\
  \Leftrightarrow&\mathbf{K'}_{u' u'}^{-1} \mathbf{S'} \mathbf{K'}_{u' u'}^{-1} = (\mathbf{K'}_{u' u'} + \mathbf{K'}_{u' u'} \mathbf{\hat{\Sigma}}^{-1} \mathbf{K'}_{u' u'})^{-1}\notag\\
  \Leftrightarrow&\mathbf{K'}_{u' u'} \mathbf{S'}^{-1} \mathbf{K'}_{u' u'} = \mathbf{K'}_{u' u'} + \mathbf{K'}_{u' u'} \mathbf{\hat{\Sigma}}^{-1} \mathbf{K'}_{u' u'}\notag\\
  \Leftrightarrow&\mathbf{K'}_{u' u'} \mathbf{S'}^{-1} \mathbf{K'}_{u' u'} - \mathbf{K'}_{u' u'} = \mathbf{K'}_{u' u'} \mathbf{\hat{\Sigma}}^{-1} \mathbf{K'}_{u' u'}\notag\\
  \Leftrightarrow&\mathbf{S'}^{-1} - \mathbf{K'}_{u' u'}^{-1} = \mathbf{\hat{\Sigma}}^{-1}\notag\\
  \Leftrightarrow& \mathbf{\hat{\Sigma}} = (\mathbf{S'}^{-1} - \mathbf{K'}_{u' u'}^{-1})^{-1},\label{ap:ssgp_Sigmahat}\\
   &\mathbf{m'} = \mathbf{K'}_{u' u'} (\mathbf{K'}_{u' u'} + \mathbf{K'}_{u' u'} \mathbf{\hat{\Sigma}}^{-1} \mathbf{K'}_{u' u'})^{-1} \mathbf{K'}_{u' u'} \mathbf{\hat{\Sigma}}^{-1} \mathbf{\hat{y}}\notag\\
  \Leftrightarrow& \mathbf{\hat{y}} = \mathbf{\hat{\Sigma}} \mathbf{K'}_{u' u'}^{-1} (\mathbf{K'}_{u' u'} + \mathbf{K'}_{u' u'} \mathbf{\hat{\Sigma}}^{-1} \mathbf{K'}_{u' u'}) \mathbf{K'}_{u' u'}^{-1} \mathbf{m'}\notag\\
  \Leftrightarrow& \mathbf{\hat{y}} = \mathbf{\hat{\Sigma}} (\mathbf{K'}_{u' u'}^{-1} + \mathbf{\hat{\Sigma}}^{-1}) \mathbf{m'}\notag\\
  \Leftrightarrow& \mathbf{\hat{y}} = \mathbf{\hat{\Sigma}} (\mathbf{K'}_{u' u'}^{-1} + \mathbf{S'}^{-1} - \mathbf{K'}_{u' u'}^{-1}) \mathbf{m'} = \mathbf{\hat{\Sigma}} \mathbf{S'}^{-1} \mathbf{m'}.\label{ap:ssgp_yhat}
\end{align}
The expressions of $\mathbf{\hat{\Sigma}}$ and $\mathbf{\hat{y}}$ are consistent with those defined in \Cref{eq:ssgp_notations} and substituting the augmented dataset and covariance into the $\mathbf{m}$ and $\mathbf{S}$ equations of SGPR in \Cref{ap:sgpr_m,ap:sgpr_S} yields the same update equations of the variational parameters $\mathbf{\tilde{m}}$ and $\mathbf{\tilde{S}}$ of SSGP.

Following the idea of constructing the pseudo data and covariance matrix by reversing some equations that involve $\mathbf{\hat{y}},\mathbf{\hat{\Sigma}}$ and some saved parameters, \citet{maddox2021conditioning} propose to cache the data-dependent terms $\color{blue} \mathbf{c}=\mathbf{K}_{u f} \mathbf{\Sigma}^{-1} \mathbf{y}$ and $\color{red} \mathbf{C}=\mathbf{K}_{u f} \mathbf{\Sigma}^{-1} \mathbf{K}_{u f}^{\T}$, and reverse these equations to obtain the pseudo targets $\mathbf{\hat{y}}$ and pseudo covariance matrix $\mathbf{\hat{\Sigma}}$, under the same assumption that the old inputs are the old inducing inputs:
\begin{align}
  &\mathbf{C'} = \mathbf{K'}_{u' u'} \bm{\hat{\Sigma}}^{-1} \mathbf{K'}_{u' u'}\notag\\
  \Leftrightarrow& \bm{\hat{\Sigma}}^{-1} = \mathbf{K'}_{u' u'}^{-1} \mathbf{C'} \mathbf{K'}_{u' u'}^{-1},\label{ap:ovc_Sigmahat}\\
                 &\mathbf{c'} = \mathbf{K'}_{u' u'} \bm{\hat{\Sigma}}^{-1} \mathbf{\hat{y}}\notag\\
  \Leftrightarrow & \mathbf{\hat{y}} = \mathbf{\hat{\Sigma}} \mathbf{K'}_{u' u'}^{-1} \mathbf{c'}\notag\\
  \Leftrightarrow & \mathbf{\hat{y}} = (\mathbf{K'}_{u' u'}^{-1} \mathbf{C'} \mathbf{K'}_{u' u'}^{-1})^{-1} \mathbf{K'}_{u' u'}^{-1} \mathbf{c'}\notag\\
  \Leftrightarrow & \mathbf{\hat{y}} = \mathbf{K'}_{u' u'} \mathbf{C'}^{-1} \mathbf{c'}.\label{ap:ovc_yhat}
\end{align}
Comparing the expressions of $\mathbf{\hat{\Sigma}}$ and $\mathbf{\hat{y}}$ in \Cref{ap:ovc_Sigmahat,ap:ovc_yhat} with those in \Cref{ap:ssgp_Sigmahat,ap:ssgp_yhat}, reversing the equations of data-dependent terms rather than the variational parameters eliminates the subtraction of two inverse matrices in the expression of $\mathbf{\hat{\Sigma}}$, which helps numerical stability.

\begin{table*}[thbp]
    \centering\scriptsize
    \renewcommand{\arraystretch}{1.5}
    \caption{\textbf{Summary of Related Work and Baseline Methods}. SSGP and OVC are two existing methods that are closely related to POAM, which are improved (i.e., SSGP++ and OVC++) to work better with non-stationary kernels and serve as strong baselines.}
    \begin{tabular}{rrrrrr}
        \toprule
        \textbf{Method} & \textbf{Saved Variables} & \textbf{Projection Matrix} & \textbf{Hyperparameters Update} & \textbf{Inducing Inputs Update}\\
        \midrule
        SSGP~\cite{bui2017streaming} & $\mathbf{S'}^{-1} \mathbf{m'}, \quad\mathbf{S'}^{-1} - \mathbf{K'}_{u' u'}^{-1}$ & $\mathbf{K}_{u' u}$ &  L-BFGS-B with Online ELBO & Resampling \& Optimization\\
        SSGP++ & $\mathbf{S'}^{-1} \mathbf{m'}, \quad\mathbf{S'}^{-1} - \mathbf{K'}_{u' u'}^{-1}$ & $\mathbf{K}_{u' u}$ &  Adam Optimizer with Online ELBO  & Pivoted Cholesky Decomposition~(PCD)\\
        OVC~\cite{maddox2021conditioning} & $\mathbf{K}_{u f} \mathbf{\Sigma}^{-1} \mathbf{y}, \quad\mathbf{K}_{u f} \mathbf{\Sigma}^{-1} \mathbf{K}_{u f}^{\T}$ & $\mathbf{K'}_{u' u'}^{-1} \mathbf{K}_{u' u}$ &  & PCD \& Optimization\\
        OVC++ & $\mathbf{K}_{u f} \mathbf{\Sigma}^{-1} \mathbf{y}, \quad\mathbf{K}_{u f} \mathbf{\Sigma}^{-1} \mathbf{K}_{u f}^{\T}$ & $\mathbf{K'}_{u' u'}^{-1} \mathbf{K}_{u' u}$ & Variational EM, Mini-Batch SGD, ELBO & PCD\\
        POAM~(ours) & $\mathbf{K'}_{u' f'} \mathbf{y'}, \quad\mathbf{K'}_{u' f'} \mathbf{K'}_{f' u'}$ & $\mathbf{K'}_{u' u'}^{-1} \mathbf{K}_{u' u}$ & Variational EM, Mini-Batch SGD, ELBO & PCD\\
        \bottomrule
    \end{tabular}
    \label{ap:literature}
\end{table*}

\citet{maddox2021conditioning} propose Online Variational Conditioning~(OVC) that computes the pseudo covariance matrix and pseudo targets following \Cref{ap:ovc_Sigmahat,ap:ovc_yhat} and feeds the augmented dataset $\{\mathbf{\tilde{X}}, \mathbf{\tilde{y}}\}$ with a pseudo likelihood function $p(\mathbf{\tilde{y}}\,|\,\mathbf{\tilde{f}})=\mathcal{N}(\mathbf{\tilde{y}} \mid \mathbf{0}, \mathbf{\tilde{\Sigma}})$ to a GPR or SGPR for online conditioning -- online update of the posterior distribution to condition on the new data.
This explains their update of variational parameters.
For inducing inputs, they use PCD on the augmented dataset. 
The target application in OVC is Bayesian optimization, where online conditioning is crucial for computing some advanced look-ahead acquisition functions while hyperparameter optimization is less important, hence not extensively discussed in their paper.

When applying the augmented dataset constructed by OVC to a SGPR, the data-dependent terms of the variational parameters are updated as
\begin{align}
  &\color{red} \mathbf{K}_{u \tilde{f}} \mathbf{\tilde{\Sigma}}^{-1} \mathbf{K}_{\tilde{f} u}\notag\\
  =& \mathbf{K}_{u f} \bm{\Sigma}^{-1} \mathbf{K}_{f u} + \mathbf{K}_{u u'} \bm{\hat{\Sigma}}^{-1} \mathbf{K}_{u' u},\notag\\
  =& {\color{red} \mathbf{K}_{u f} \bm{\Sigma}^{-1} \mathbf{K}_{f u}} + \mathbf{K}_{u' u}^{\T} ({\color{green}\mathbf{K'}_{u' u'}^{-1} \mathbf{C'} \mathbf{K'}_{u' u'}^{-1}}) \mathbf{K}_{u' u},\label{ap:ovc_KK}\\
   &\color{blue} \mathbf{K}_{u \tilde{f}} \mathbf{\tilde{\Sigma}}^{-1} \mathbf{\tilde{y}}\notag\\
  =& \mathbf{K}_{u f} \bm{\Sigma}^{-1} \mathbf{y} + \mathbf{K}_{u u'} \bm{\hat{\Sigma}}^{-1} \mathbf{\hat{y}},\notag\\
  =& \mathbf{K}_{u f} \bm{\Sigma}^{-1} \mathbf{y} + \mathbf{K}_{u' u}^{\T} (\mathbf{K'}_{u' u'}^{-1} \mathbf{C'} \mathbf{K'}_{u' u'}^{-1}) \mathbf{K'}_{u' u'} \mathbf{C'}^{-1} \mathbf{c'},\notag\\
  =& {\color{blue} \mathbf{K}_{u f} \bm{\Sigma}^{-1} \mathbf{y}} + \mathbf{K}_{u' u}^{\T} {\color{green}\mathbf{K'}_{u' u'}^{-1} \mathbf{c'}}\label{ap:ovc_Ky}.
\end{align}
\Cref{ap:ovc_KK,ap:ovc_Ky} can be viewed as updating $\mathbf{C'}$ and $\mathbf{c'}$ via a projection matrix $\mathbf{P}\triangleq \mathbf{K'}_{u' u'}^{-1} \mathbf{K}_{u' u}$ and adding the new data-dependent terms:
\begin{align}
  \mathbf{C} =& \mathbf{K}_{u f} \bm{\Sigma}^{-1} \mathbf{K}_{f u} + \mathbf{P}^{\T} \mathbf{C'} \mathbf{P},\label{ap:ovc_upper_c}\\
  \mathbf{c} =& \mathbf{K}_{u f} \bm{\Sigma}^{-1} \mathbf{y} + \mathbf{P}^{\T} \mathbf{c'}.\label{ap:ovc_lower_c}
\end{align}

Our proposed POAM shares the same idea of projecting the old data-dependent terms and adding the new data-dependent terms, but we save $\mathbf{K'}_{u' f'} \mathbf{y'}$ and $\mathbf{K'}_{u' f'} \mathbf{K'}_{f' u'}$ instead of $\mathbf{c'}$ and $\mathbf{C'}$, which leads to better performance in our experiments.
In other words, the observational noise $\mathbf{\Sigma}=\sigma^{2}\mathbf{I}$ is not data-dependent, so we should use the new $\mathbf{\Sigma}$ rather than projecting the old one.

\subsubsection{Baseline Methods}
We found that optimizing the inducing inputs with the online ELBO of SSGP performs poorly in our experiments, regardless of the initialization strategy, so SSGP++ uses pivoted Cholesky decomposition~(PCD)~\cite{burt2020convergence} for updating inducing inputs and does not optimize them with the online ELBO.
Also, SSGP++ uses the Adam optimizer for updating the hyperparameters because the L-BFGS-B optimizer is not suitable for the Attentive Kernel~(AK) that has many more parameters than the commonly used stationary kernels.
The hyperparameter optimization of OVC is not properly discussed in \cite{maddox2021conditioning}, so we use the same strategy as POAM.
The vanilla OVC still optimizes the inducing inputs after initializing them with PCD, while OVC++ does not optimize them.
\Cref{ap:literature} summarizes the differences of each component in the aforementioned methods.

We note that during the experiments, SSGP++ sometimes throws a numerical error, requiring an increase in the ``jitter'' value added to the diagonal elements of positive-definite matrices for stable Cholesky decomposition and re-running the experiment. Although not reported in the results, this is an important factor to consider when deploying SSGP++ in practice.



\begin{figure*}[htbp]
  \centering
  \subfloat{{\includegraphics[width=0.7\linewidth]{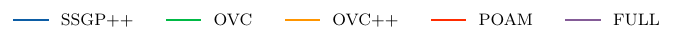}}}\vspace{-1em}\\
  \addtocounter{subfigure}{-1}\vspace{1em}\textbf{Random Initialization}\\
  \subfloat[SMSE in Env1]{\includegraphics[width=0.25\linewidth,height=0.13\linewidth,trim={0 4 0 1},clip]{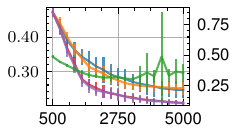}}%
  \subfloat[SMSE in Env2]{\includegraphics[width=0.25\linewidth,height=0.13\linewidth,trim={0 4 0 1},clip]{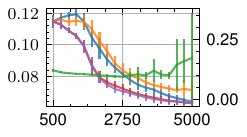}}%
  \subfloat[SMSE in Env3]{\includegraphics[width=0.25\linewidth,height=0.13\linewidth,trim={0 4 0 1},clip]{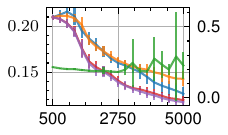}}%
  \subfloat[SMSE in Env4]{\includegraphics[width=0.25\linewidth,height=0.13\linewidth,trim={0 4 0 1},clip]{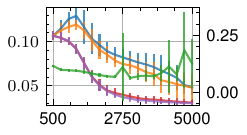}}\\%
  \subfloat[MSLL in Env1]{\includegraphics[width=0.24\linewidth,height=0.13\linewidth,trim={0 4 0 1},clip]{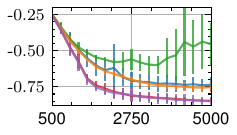}}%
  \subfloat[MSLL in Env2]{\includegraphics[width=0.24\linewidth,height=0.13\linewidth,trim={0 4 0 1},clip]{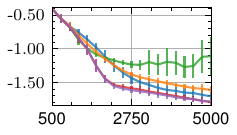}}%
  \subfloat[MSLL in Env3]{\includegraphics[width=0.24\linewidth,height=0.13\linewidth,trim={0 4 0 1},clip]{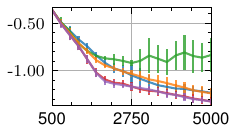}}%
  \subfloat[MSLL in Env4]{\includegraphics[width=0.24\linewidth,height=0.13\linewidth,trim={0 4 0 1},clip]{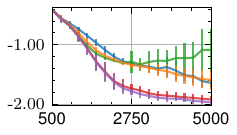}}\\%
  \vspace{1em}\textbf{Random Planner}\\
  \subfloat[SMSE in Env1]{\includegraphics[width=0.24\linewidth,height=0.13\linewidth,trim={0 4 0 1},clip]{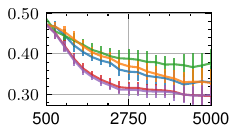}}%
  \subfloat[SMSE in Env2]{\includegraphics[width=0.24\linewidth,height=0.13\linewidth,trim={0 4 0 1},clip]{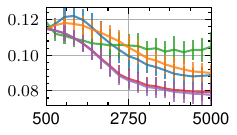}}%
  \subfloat[SMSE in Env3]{\includegraphics[width=0.24\linewidth,height=0.13\linewidth,trim={0 4 0 1},clip]{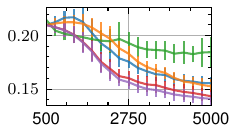}}%
  \subfloat[SMSE in Env4]{\includegraphics[width=0.24\linewidth,height=0.13\linewidth,trim={0 4 0 1},clip]{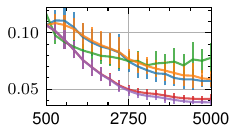}}\\%
  \subfloat[MSLL in Env1]{\includegraphics[width=0.24\linewidth,height=0.13\linewidth,trim={0 4 0 1},clip]{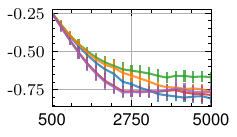}}%
  \subfloat[MSLL in Env2]{\includegraphics[width=0.24\linewidth,height=0.13\linewidth,trim={0 4 0 1},clip]{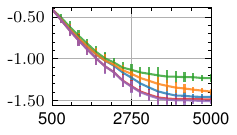}}%
  \subfloat[MSLL in Env3]{\includegraphics[width=0.24\linewidth,height=0.13\linewidth,trim={0 4 0 1},clip]{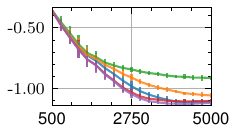}}%
  \subfloat[MSLL in Env4]{\includegraphics[width=0.24\linewidth,height=0.13\linewidth,trim={0 4 0 1},clip]{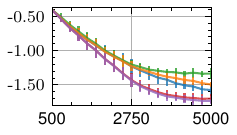}}%
  \caption{Standardized Mean Squared Error (SMSE) and Mean standardized log-loss (MSLL) curves for three baseline methods (SSGP++, OVC, OVC++) are compared against the proposed method (POAM) and its full-update version (FULL) which performs a full update of the variational parameters at each time step. The random initialization experiments use the max-entropy planner and the initial samples and robot's starting positions are uniformly and randomly sampled from the environment. The random planner experiments use the same random-initialization strategy and a random planner that uniformly samples a waypoint at random from the environment at each decision epoch. Note that, in the random initialization experiments, the y-axis of SMSE on the right is for the OVC method alone, as its performance is much worse than the other approaches.}\label{fig:rebuttal}%
\end{figure*}

\balance

\subsection{Additional Experiments}%
To gauge the robustness of the proposed method to different initial conditions and to disentangle the contributions of the proposed model and the planner, we have conducted two additional sets of experiments.
These experiments also include an additional baseline as follows.
\begin{itemize}
  \item \textbf{Random Initialization}: A set of experiments with initial samples and the robot's starting positions \emph{uniformly and randomly} sampled from the environment and the same \emph{max-entropy planner} used in the original experiments. These experiments are conducted to evaluate the robustness of the proposed method to different initial conditions.
  \item \textbf{Random Planner}: A set of experiments with the same random-initialization strategy and a \emph{random planner} that uniformly samples a waypoint at random from the environment at each decision epoch. These experiments compare different models when the robot executes the same path that is independent of the model being evaluated.
  \item \textbf{Full dataset baseline}:
  We include an additional baseline using
the full-update version of POAM (\texttt{FULL}) to verify the effectiveness of the proposed online update rule. The \texttt{FULL} method uses the full dataset to update the variational parameters instead of using the proposed online update rule, so it is expected to be slower but has better performance compared to \texttt{POAM}.
\end{itemize}

 The experiments are repeated $10$ times with different random seeds and the mean and standard deviation of the results are reported in \Cref{fig:rebuttal}.
 
 \topic{Results of Random Initialization Experiments}
 Under the randomized initialization, the rankings of the compared methods are consistent across different environments: the proposed method (\texttt{POAM}) and its full-update version (\texttt{FULL}) outperform the baselines in terms of SMSE and MSLL, \texttt{SGGP++} and \texttt{OVC++} perform similarly, and \texttt{OVC} performs the worst. Since the SMSE of OVC is much worse than other approaches in the random initialization experiments, we use a separate y-axis at the right side of the SMSE plots to better visualize the performance of the other methods.

 \topic{Results of Random Planner Experiments}
 \texttt{POAM} outperforms the baselines in terms of SMSE and MSLL, \texttt{SGGP++} and \texttt{OVC++} perform similarly, and OVC performs the worst. Additionally, by contrasting the two sets of experiments which only differ in the planner, we can see the performance gain brought by the active learning strategy is also evident.

 \topic{Comparison with Full Update}
The difference between the proposed online update rule and the full update is negligible. This indicates that the proposed online update rule is accurate and robust, without significant error accumulation over time.

\end{document}